\documentclass[twoside,11pt]{article}
\usepackage{jair, theapa, rawfonts}

\usepackage{graphicx}
\usepackage[cmex10]{amsmath}
\usepackage{amsfonts}
\usepackage{amsmath}
\usepackage{amsthm}
\usepackage{algorithmic}
\usepackage{array}
\usepackage{mdwmath}
\usepackage{mdwtab}
\usepackage{eqparbox}
\usepackage{color}
\usepackage{algorithm}
\usepackage{url}

\newtheorem{proposition}{Proposition}

\ShortHeadings{Combinatorial Multi-armed Bandits for Real-Time Strategy Games}
{Onta\~{n}\'{o}n}

\begin{document}

\title{Combinatorial Multi-armed Bandits \\ for Real-Time Strategy Games}

\author{\name Santiago~Onta\~{n}\'{o}n \email santi@cs.drexel.edu \\
		\addr Computer Science Department, Drexel University \\
		Philadelphia, PA, USA}

\maketitle

\begin{abstract}
Games with large branching factors pose a significant challenge for game tree search algorithms. In this paper, we address this problem with a  sampling strategy for Monte Carlo Tree Search (MCTS) algorithms called {\em na\"{i}ve sampling}, based on a variant of the Multi-armed Bandit problem called {\em Combinatorial Multi-armed Bandits} (CMAB). We analyze the theoretical properties of several variants of {\em na\"{i}ve sampling}, and empirically compare it against the other existing strategies in the literature for CMABs. We then evaluate these strategies in the context of real-time strategy (RTS) games, a genre of computer games characterized by their very large branching factors. Our results show that as the branching factor grows, {\em na\"{i}ve sampling} outperforms the other sampling strategies.
\end{abstract}

\section{Introduction}\label{sec:intro}

Games with large branching factors pose a significant challenge for game tree search algorithms. So far, Monte Carlo Tree Search (MCTS) algorithms \cite{browne2012survey}, such as UCT \cite{Kocsis06banditbased}, are the most successful approaches for this problem. The key to the success of MCTS algorithms is that they sample the search space, rather than exploring it systematically. However, MCTS algorithms quickly reach their limit when the branching factor grows. To illustrate this, consider Real-Time Strategy (RTS) games,
where each player controls a collection of units, all of which can be controlled simultaneously, leading to a combinatorial branching factor. For example, just 10 units with 5 actions each results in a potential branching factor of $5^{10} \approx 10$ million, beyond what standard MCTS algorithms can handle. Algorithms that can handle adversarial planning in situations with combinatorial branching factors would have many applications to problems such as multiagent planning.

Specifically, this paper focuses on scaling up MCTS algorithms to games with combinatorial branching factors. MCTS algorithms formulate the problem of deciding which parts of the game tree to explore as a Multi-armed Bandit (MAB) problem \cite{auer2002finite}. In this paper, we will show that by considering a variant of the MAB problem called the {\em Combinatorial Multi-armed Bandit} (CMAB) \cite{gai2010learning,chen2013combinatorial,ontanon2013combinatorial}, it is possible to handle the larger branching factors appearing in RTS games.

Building on our previous work in this area \cite{ontanon2013combinatorial}, where we first introduced the idea of {\em na\"{i}ve sampling}, the main contributions of this paper are: (1) an analysis of the different instantiations of the family of {\em na\"{i}ve sampling} strategies, including regret bounds; (2) an empirical comparison with other existing CMAB sampling strategies in the literature (LSI, see \citeR{shleyfman2014combinatorial}; and MLPS, see \citeR{gai2010learning}); (3) empirical results using increasingly complex situations, to understand the performance of these strategies as the problems grow in size (reaching situations with branching factors in the order of $10^{22}$).

We use the $\mu$RTS game simulator\footnote{\url{https://github.com/santiontanon/microrts}} as our application domain, which is a deterministic and fully-observable RTS game (although it can be configured for partial observability or non-determinism). Our results indicate that for scenarios with small branching factors, {\em na\"{i}ve sampling} performs similar to other sampling strategies, but as the branching factor grows, {\em na\"{i}ve sampling} starts outperforming the other approaches. A snapshot of all the source code and data necessary to reproduce all the experiments presented in this paper can be downloaded from the author's website\footnote{\url{https://sites.google.com/site/santiagoontanonvillar/code/NaiveSampling-journal-2016-} \url{source-code.zip}}. 

The remainder of this paper is organized as follows. Section \ref{sec:background} presents some background on RTS games and MCTS. Section \ref{sec:cmab} then introduces the CMAB problem. Section \ref{sec:naive} introduces and analyzes na\"{i}ve sampling strategies, after which Section \ref{sec:other-cmab-strategies} presents other known sampling strategies for CMABs in the literature. All of these strategies are compared empirically in Section \ref{sec:sampling-comparison}. After that, we describe how to integrate them into MCTS in Section~\ref{sec:mcts}, and the strength of the resulting MCTS algorithm is evaluated empirically in the $\mu$RTS simulator in Section \ref{sec:experiments}. The paper closes with related work, conclusions, and directions for future research.

\section{Background}\label{sec:background}

The following two subsections present some background on real-time strategy (RTS) games, and on Monte Carlo Tree Search in the context of RTS games.

\subsection{Real-Time Strategy Games}

Real-time Strategy (RTS) games are complex adversarial domains, typically simulating battles between a large number of military units, that pose a significant challenge to both human and artificial intelligence \cite{buro2003rts}. Designing AI techniques for RTS games is challenging because:
\begin{itemize}
\item They have {\em huge decision and state spaces}: To have a sense of scale, the worst case branching factor of a typical RTS game, StarCraft, has been estimated to be at least $10^{50}$ \cite{starcraft2013survey} when the player can control all units simultaneously, which is staggering if we compare it with the branching factors of games like Chess (about 36) and Go (about 180). Moreover, the state space of StarCraft has been estimated to be at least $10^{1685}$ \cite{starcraft2013survey}, compared to about $10^{47}$\cite{Chinchalkar1996upperbound} for Chess and $10^{171}$\cite{tromp2006combinatorics} for Go.

\item They are {\em real-time}, which means that: (1) RTS games typically execute at 10 to 50 decision cycles per second, leaving players with just a fraction of a second to decide the next move; (2) players do not take turns, but can issue actions simultaneously (i.e., two players can issue actions at the same instant of time, and to as many units as they want); and (3) actions are durative, i.e., actions might take more than one decision cycle to complete.
\end{itemize}
Some RTS games are also partially observable and non-deterministic, but we will not deal with these properties in this paper.

In RTS games, players control a collection of individual units that players issue actions to. Each of these units can only execute one action at a time, but, since there might be multiple units in a game state, players can issue multiple actions at the same time (one per unit they control). We will refer to those actions as {\em unit-actions}, and use lower case $a$ to denote them. A {\em player-action} $\alpha$ is the set of unit-actions that one player issues simultaneously at a given time: $\alpha = \{a_1, ..., a_n\}$. Thus, without loss of generality, we can consider that players issue only one {\em player-action} per game decision cycle (which will consist of as many unit-actions as units ready to execute an action in the current decision cycle). In this way, even if unit-actions are durative, we can see an RTS game as a game where each player issues exactly one player-action at each decision cycle. The number of possible player-actions corresponds to the branching factor. Thus, the branching factor in a RTS game grows exponentially with the number of units each player controls (without loss of generality, we can assume a special {\em no-op} unit-action, to be issued to those units the player does not want to do anything in the current decision cycle).

\begin{figure}[t]
    \centering
    \includegraphics[width=0.45\columnwidth]{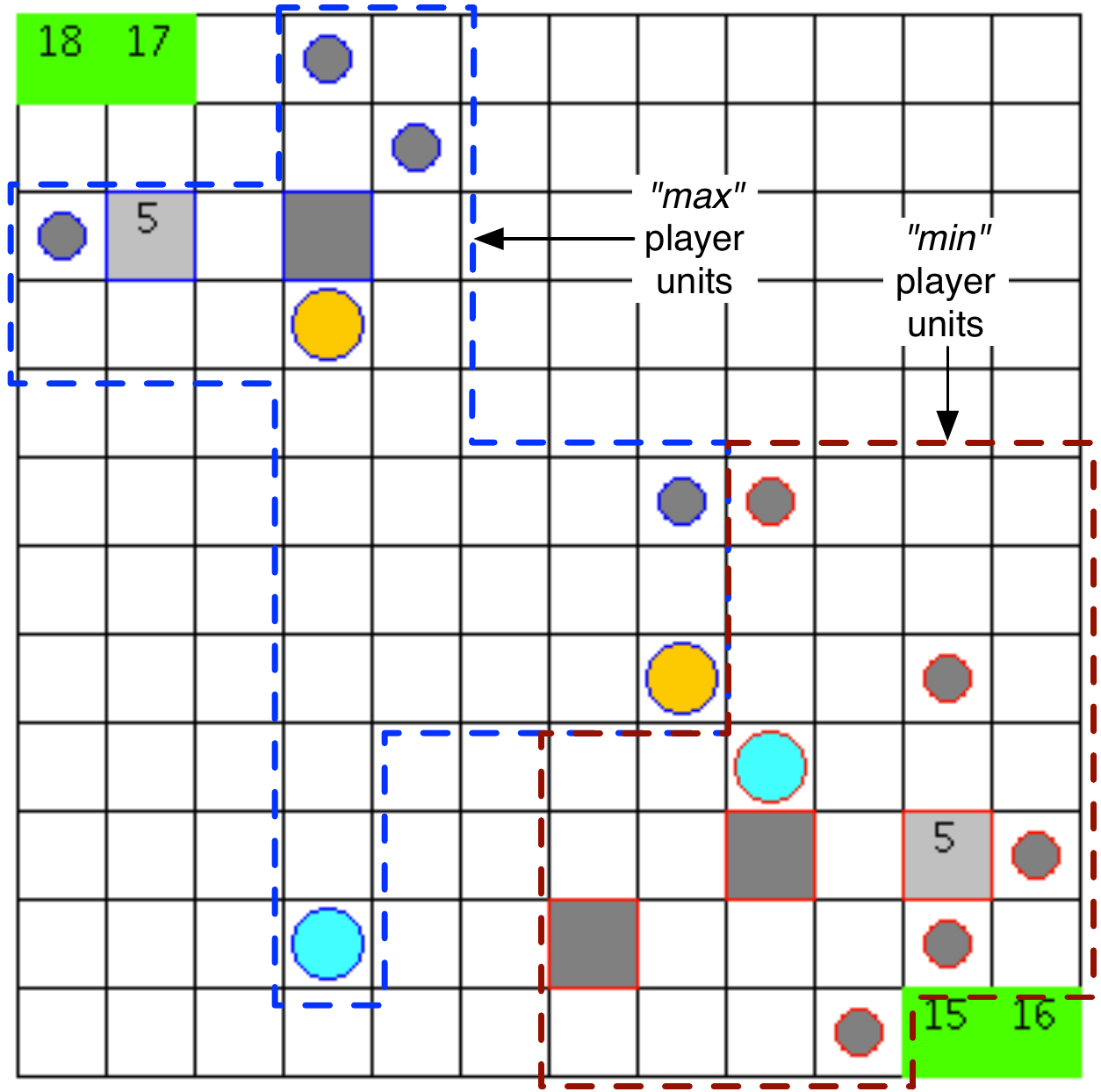}
    \caption{A screenshot of the $\mu$RTS simulator. Square units correspond to ``bases'' (light grey, that can produce workers), ``barracks'' (dark grey, that can produce military units), and ``resources mines'' (green, from where workers can extract resources to produce more units), the circular units correspond to workers (small, dark grey) and military units (large, yellow or light blue).}
    \label{fig:microRTS}
\end{figure}

To illustrate the size of the branching factor in RTS games, consider the situation from the $\mu$RTS game\footnote{For gaining a more intuitive idea of $\mu$RTS, a gameplay video can be found here: \url{https://www.youtube.com/watch?v=Or3IZaRRYIQ}} (used in our experiments) shown in Figure \ref{fig:microRTS}. Two players, {\em max} (shown in blue) and {\em min} (shown in red) control 9 units each. Consider the bottom-most circular unit in Figure \ref{fig:microRTS} (a worker). This unit can execute 8 actions: stand still, move left or up, harvest the resource mine to the right, or build a barracks or a base in any of the two free adjacent cells. In total, player {\em max} in Figure \ref{fig:microRTS} can issue 1,008,288 different player-actions, and player {\em min} can issue 1,680,550 different player-actions. Thus, even in relatively simple scenarios, the branching factor is very large. 

Specifically, a two-player deterministic perfect-information RTS game is a tuple $G = (S, A, P, \tau, L, W, s_{init})$, where:

\begin{itemize}
\item $S$ is the game state space (e.g., in Chess, the set of all possible board configurations).
\item $A$ is the finite set of possible player-actions that can be executed in the game. 
\item $P = \{max,min\}$ is the set of players.
\item $\tau : S \times A \times A \to S$ is the deterministic transition function, that given a state at time $t$, and the actions of the two players, returns the state at time $t+1$.
\item $L : S \times A \times P \to \{true,false\}$ is a function that given a state, a player-action and a player, determines whether it is legal to execute the given player-action by the given player in the given state. We will write $actions(s,p) = \{ \alpha \in A | L(s,\alpha,p) = true\}$ to denote the set of player-actions that player $p$ can execute in state $s$.
\item $W : S \to P \cup \{draw, ongoing\}$ is a function that given a state determines the winner of the game, if the game is still ongoing, or if it is a draw.
\item $s_{init} \in S$ is the initial state.
\end{itemize}

In order to apply game tree search, an additional {\em evaluation function} is typically provided. The evaluation function predicts how attractive is a given state for a player. We will assume an evaluation function of the form $\rho : S \to \mathbb{R}$, which returns positive numbers for states that are good for $max$ and negative numbers for states that are good for $min$.

\subsection{Monte Carlo Tree Search in RTS Games}\label{subsec:mcts}

Monte Carlo Tree Search (MCTS) is a family of planning algorithms based on sampling the decision space rather than exploring it systematically \cite{browne2012survey}. 
MCTS algorithms maintain a partial game tree. Each node in the tree corresponds to a game state, and the children of that node correspond to the result of one particular player executing actions. Additionally, each node stores the number of times it has been explored, and the average reward obtained when exploring it. Initially, the tree contains a single root node with the initial state. Then, assuming the existence of a reward function $\rho$, at each iteration of the algorithm the following three processes are executed:
\begin{itemize}
\item {\em SelectAndExpandNode}: Starting from the root node, one of the current node's children is chosen following a {\em tree policy}, until a node $n$ that was not in the tree before is reached. The new node $n$ is added to the tree.
\item {\em Simulation}: A Monte Carlo {\em simulation}  (a.k.a. a {\em playout} or a {\em rollout}) is executed starting from $n$ using a {\em default policy} (e.g., random) to select actions for all the players until a terminal state or a maximum simulation time is reached. Let $r = \rho(s)$ be the reward in the state $s$ at the end of the simulation.
\item {\em Backup}: $r$ is propagated up the tree, starting from the node $n$, and continuing through all the ancestors of $n$ in the tree (updating their average reward, and incrementing by one the number of times they have been explored). 
\end{itemize}
When the computation budget is over, the action that leads to the ``best'' child of the root node of the tree is selected as the best action to perform. Here, ``best'' can be defined as the one with highest average reward, the most visited one, or some other criteria (depending on the tree policy). 

Different MCTS algorithms typically differ just in the tree policy. In particular, UCT \cite{Kocsis06banditbased} frames the tree policy as a {\em Multi-armed Bandit} (MAB) problem.
MAB problems are a class of sequential decision problems, where at each iteration an agent needs to choose amongst $k$ actions (called {\em arms}), in order to maximize the cumulative reward obtained by those actions. A MAB problem with $k$ arms is defined by a set of unknown real reward distributions $\mu_1, ..., \mu_k$, associated with each of the $k$ arms. Therefore, the agent needs to estimate the potential rewards of each action based on past observations, balancing exploration and exploitation.

UCT uses a specific sampling strategy called UCB1 \cite{auer2002finite} that balances exploration and exploitation of the different nodes in the tree. It can be shown that, when the number of iterations executed by UCT approaches infinity, the probability of selecting a suboptimal action approaches zero \cite{Kocsis06banditbased}. However, UCB1 does not scale well to the domains of interest in this paper, where the branching factor might be several orders of magnitude larger than the number of samples we can perform. 

Previous work has addressed many of the key challenges arising in applying game tree search to RTS games. For example game tree search algorithms exist that can handle durative actions \cite{churchill2012abcd}, or simultaneous moves \cite{KovarskyB05heuristic,SaffidineFinnssonBuro2012AAAI}. However, the branching factor in RTS games remains too large for current state-of-the-art techniques. 

Many ideas have been explored to improve UCT in domains with large branching factors. For example, {\em first play urgency} (FPU) \cite{gelly2006exploration} allows the bandit strategy of UCT (UCB) to exploit nodes early, instead of having to visit all of them before it starts exploiting. However, FPU still does not address the problem of selecting which of the unexplored nodes to explore first (which is key in our domains of interest). Another idea is to try to maximally exploit the information obtained from each simulation, like performed by AMAF \cite{gelly2007combining}. However, again, this does not solve the problem of having a branching factor many orders of magnitude larger than the number of simulations we can perform. 
As elaborated in Section \ref{sec:related}, three main approaches have been explored to address this problem: (1) use abstraction to represent the game state or the action space to simplify the problem \cite{balla2009uct,uriarte2014game}; (2) portfolio approaches that only consider moves chosen by a predefined portfolio of strategies \cite{churchill2013portfolio,ChungBS05}; or (3) hierarchical search approaches that aim at pruning the search space by considering first high-level decisions, which condition the potential number of low-level decisions that can be taken \cite{stanescu2014hierarchical,ontanon2015adversarial}. This paper studies an alternative idea, namely using {\em combinatorial multi-armed bandits} (CMAB), which have recently been proposed as a solution to address the combinatorial branching factors arising in RTS games \cite{ontanon2013combinatorial,shleyfman2014combinatorial}.

\section{Combinatorial Multi-armed Bandits}\label{sec:cmab}

A  {\em Combinatorial Multi-armed Bandit} (CMAB) is a variation of the MAB problem. We use the formulation in \citeauthor{ontanon2013combinatorial}~\citeyear{ontanon2013combinatorial}, which is more general than that by \citeauthor{gai2010learning}~\citeyear{gai2010learning} or that by \citeauthor{chen2013combinatorial}~\citeyear{chen2013combinatorial}. Specifically, a CMAB is defined by:
\begin{itemize}
\item A set of $n$ variables $X = \{X_1, ..., X_n\}$, where variable $X_i$ can take $K_i$ different values $\mathcal{X}_i = \{v^1_i, ..., v^{K_i}_i\}$, and each of those values is called an {\em arm}. Let us call $\mathcal{X} = \{ (v_1, ..., v_n) \in \mathcal{X}_1 \times ... \times \mathcal{X}_n\}$ to the set of possible value combinations, where each of these combinations $V \in \mathcal{X}$ is called a {\em macro-arm}.
\item An unknown reward distribution $\mu : \mathcal{X} \to \mathbb{R}$ over each macro-arm. 
\item A function $L : \mathcal{X} \to \{true,false\}$ that determines which macro-arms are legal.
\end{itemize}
The problem is to find a legal macro-arm that maximizes the expected reward. Strategies to address CMABs are designed to iteratively sample the space of possible macro-arms. At each iteration $t$, one macro-arm $V_t$ is selected, which results in a given reward  $\mu_t = \mu(V_t)$. Strategies must balance exploration and exploitation in order to converge to the best macro-arm in the shortest number of iterations possible. We will call $V_t^{best}$ to the macro-arm recommended as the best by the given strategy after iteration $t$.

The difference between a MAB and a CMAB is that in a MAB there is a single variable, whereas in a CMAB, there are $n$ variables. A CMAB can be translated to a MAB, by considering that each possible legal macro-arm is a different arm in the MAB\@. Moreover, the possible number of macro-arms in a CMAB grows exponentially with the number of variables (depending on how many of those macro-arms are legal).

The performance of strategies to address MABs and CMABs is assessed by measuring the {\em regret}, which is the difference between the expected reward of the selected macro-arm, and the expected reward of an optimal macro-arm. Moreover, regret can be computed in several different ways \cite{bubeck2011pure}, assuming that $V^*$ is an optimal macro-arm, obtaining maximum expected reward $\mu^* = E(\mu(V^*))$:
\begin{itemize}
\item {\em Instantaneous regret}: is the difference between $\mu^*$ and the reward obtained by the last selected macro-arm. After $T$ iterations, the instantaneous regret is computed as:
$$r'_T = \mu^* - \mu_T$$

\item {\em Cumulative regret} (referred to as {\em pseudo-regret} in some texts, see \citeR{bubeck2012regret}): is the sum of differences between $\mu^*$ and the reward obtained by the selected macro-arms at each iteration. After $T$ iterations, the cumulative regret is:
$$R_T = \sum_{t=1}^T (\mu^* - \mu_t)$$
\noindent where $\mu_t$ is the reward obtained at iteration $t$.

\item {\em Simple regret}: after $T$ iterations, the simple regret is the difference between $\mu^*$ and the reward obtained by the macro-arm believed to be the best at iteration $T$:
$$r_T = \mu^* - \mu(V_T^{best})$$
\end{itemize}
Thus, instantaneous regret is the difference in reward at a given time $t$ based on the selected macro-arm, cumulative regret is the sum of all the instantaneous regrets so far, and simple regret is the instantaneous regret that would be obtained if the arm recommended as the best at iteration $t$ is chosen. 

As pointed out by \citeauthor{bubeck2011pure}~\citeyear{bubeck2011pure}, strategies that minimize cumulative regret, obtain larger simple regrets, and vice-versa. Therefore, it is important to determine which kind of regret we must minimize in the task we are modeling using MABs or CMABs.  As pointed out by Tolpin and Shimony \citeyear{tolpin2012mcts}, and by \citeauthor{shleyfman2014combinatorial}~\citeyear{shleyfman2014combinatorial}, bandit strategies applied to planning in games should minimize simple regret, since the performance of the agent in the game is based only on the performance of the final action selected, which corresponds to simple regret. Thus, it might seem that standard approaches to MCTS, such as UCT which uses UCB1 \cite{auer2002finite} and thus minimize cumulative regret, are minimizing the wrong measure. However, notice that  bandit strategies running in the search nodes of an MCTS algorithm need to balance two main objectives: (1) identify the best action, and (2) estimate the reward of the best action. While the first one is achieved by doing pure exploration (aiming at minimizing simple regret), the later is not. Thus, bandit strategies in MCTS algorithms need to strike a balance between exploration and exploitation in order to achieve both objectives. Recently, however, several MCTS algorithms have been designed to directly minimize simple regret. Examples are BRUE \cite{feldman2014simple}, MCTS SR+CR \cite{tolpin2012mcts}, or SHOT \cite{cazenave2015sequential}. For simplicity, however, in our experimental evaluation, we will use a standard MCTS algorithm. 

In this paper, we will use CMABs to model the decision process that a player faces in RTS games. Each of the units in the game state will be modeled with a variable $X_i$, and the values that each of these variables can take correspond to the unit-actions that the corresponding units can execute. Player-actions thus naturally correspond to macro-arms.

\section{Na\"{i}ve Sampling For CMABs}\label{sec:naive}

{\em Na\"{i}ve sampling} (NS) is a family of sampling strategies based on assuming that the reward distribution $\mu$ can be approximated as the sum of a set of reward functions $\mu_1, ..., \mu_n$, each of them depending only on the value of one of the variables of the CMAB:
$$\mu(X) \approx \sum_{i = 1 ... n} \mu_i(X_i)$$ 
We call this the {\em na\"{i}ve assumption}, since it is reminiscent of the conditional independence assumption of the Na\"{i}ve Bayes classifier. Thanks to the na\"{i}ve assumption, we can break the CMAB problem into a collection of $n+1$ MAB problems. 

\begin{itemize}
\item {\em Local MABs}: For each $X_i \in X$, we define a MAB, $\mathit{MAB}_i$, that only considers $X_i$. 
\item {\em Global MAB}: $\mathit{MAB}_g$, that considers the whole CMAB problem as a MAB where each legal macro-arm that has been sampled so far is one of the candidate arms. This means that in the first iteration, $t=1$, the global MAB contains no arms at all.
\end{itemize}

Intuitively, na\"{i}ve sampling uses the local MABs to {\em explore} different macro-arms that are likely to result on a high reward, and then uses the global MAB to {\em exploit} the macro-arms that obtained the highest reward so far. Let us first introduce some notation:

\begin{itemize}
\item Let $T^t_i(v_i^k)$ be the number of times that value $v_i^k$ has been selected for variable $X_i$ up to iteration $t$.
\item Let $\overline{\mu}^t_i(v_i^k)$ be the marginalized average reward obtained when selecting value $v_i^k$ for variable $X_i$ up to time $t$.
\item Let $T^t(v_1^{k_1}, ..., v_n^{k_n})$ be the number of times that macro-arm $(v_1^{k_1}, ..., v_n^{k_n})$ has been selected up to time $t$.
\item Let $\overline{\mu}^t(v_1^{k_1}, ..., v_n^{k_n})$ be the average reward obtained when selecting the macro-arm $(v_1^{k_1}, ..., v_n^{k_n})$ up to time $t$.
\end{itemize}

The NS strategy works as follows. At each iteration $t$:

\begin{itemize}
\item Use a strategy $\pi_0$ to determine whether to {\em explore} (via the local MABs) or {\em exploit} (via the global MAB). 

\begin{itemize}
\item If {\em explore} was selected: a legal macro-arm $x^t = (x_1^t, ..., x_n^t)$ is selected by using a strategy $\pi_l$ to select a value for each $X_i \in X$ independently (i.e., the strategy is used $n$ times, one per variable). $x^t$ is added to the global MAB\@. 
\item If {\em exploit} was selected: a macro-arm $x^t$ is selected by using a strategy $\pi_g$ over the macro-arms already present in  the global MAB\@. 
\end{itemize}
\end{itemize}

Intuitively, when exploring, the na\"{i}ve assumption is used to select values for each variable, assuming that this can be done independently using the estimated $\overline{\mu}^t_i$ expected rewards. At each iteration, the selected macro-arm is added to the global MAB, $\mathit{MAB}_g$. Since the assumption is that the number of iterations $T$ that we can perform is much smaller than $N$ (the total number of macro-arms), it is expected that almost each time that the strategy decides to explore, the selected macro-arm is not going to be already in the global MAB\@.

When exploiting, $\mathit{MAB}_g$ is used to sample amongst the explored macro-arms, and find the one with the expected maximum reward. Thus, we can see that the na\"{i}ve assumption is used to explore the combinatorial space of possible macro-arms, and then a regular MAB strategy is used over the global MAB to select the optimal macro-arm. If the strategy $\pi_l$ is selected such that each arm has a non-zero probability of being selected, then each possible value combination also has a non-zero probability. Thus, the error in the estimation of $\overline{\mu}^t$ constantly decreases. As a consequence, the optimal value combination will eventually have the highest estimated reward, regardless of whether the reward function violates the na\"{i}ve assumption or not. Thus, notice that in order to work well, na\"{i}ve sampling only requires the game domain to satisfy the na\"{i}ve assumption loosely (i.e., that macro-arms composed of unit-actions that have individually high reward, also tend to have a high reward).

Moreover, NS is not just one sampling strategy, but a whole family of sampling strategies, since we still need to decide which sampling strategies to use for $\pi_0$, $\pi_l$, and $\pi_g$. In our previous work, we studied the performance when all three strategies are $\epsilon$-greedy strategies \cite{ontanon2013combinatorial}. Let us now analyze the behavior of NS for different instantiations of these strategies. The following subsections first present the theoretical regret bounds, and then an empirical comparison of the performance of these strategies in the $\mu$RTS simulator. Evaluation of the performance of these strategies in the context of game tree search is presented in Section \ref{sec:experiments}.

\subsection{$\epsilon$-greedy Na\"{i}ve Sampling}

One of the most common MAB sampling strategies is $\epsilon$-greedy. An $\epsilon$-greedy strategy with parameter $0 \leq \epsilon \leq 1$ selects the arm considered to be the best one so far with probability $1-\epsilon$, and with probability $\epsilon$ it selects one arm at random. If $\epsilon$ is small (e.g., 0.1), this results on a behavior that selects the arm currently considered the best most of the times, but keeps exploring all the other arms with a small probability.

We will write NS($\epsilon_0$,$\epsilon_l$,$\epsilon_g$) to denote a na\"{i}ve sampling where $\pi_0$, $\pi_l$, and $\pi_g$ are $\epsilon$-greedy strategies, with parameters $0 \leq \epsilon_0 \leq 1$ ($\epsilon_0$ probability of selecting {\em explore} and $1-\epsilon_0$ of selecting {\em exploit}), $0 \leq \epsilon_l \leq 1$, and $0 \leq \epsilon_g \leq 1$ respectively. Let us now see how the regret of this strategy grows over time.

\begin{proposition}\label{ens-cumulative-proposition}
The {\em cumulative regret} of NS($\epsilon_0$,$\epsilon_l$,$\epsilon_g$) grows linearly as $R_T = O((1 - p^*) D T)$, where $T$ is the number of iterations, $D$ is the expected difference in expected reward between an optimal macro-arm and a non-optimal macro-arm, and $p^* \geq (1 - \epsilon_0)(1 - \epsilon_g)$ is the probability of selecting an optimal arm when $T \to \infty$ 
(proof in Appendix \ref{sec:appendix}).
\end{proposition}

Moreover, if we assume a single optimal arm and we know that the reward function $\mu$ satisfies the na\"{i}ve assumption, we can be more precise, and get an exact value for $p^*$:
$$ p^* = (1-\epsilon_0)\left[(1-\epsilon_g) + \frac{\epsilon_g}{N} \right] + \epsilon_0 \prod_{i = 1 ... n}\left[(1-\epsilon_l) + \frac{\epsilon_l}{K_i}\right] $$
Where $N$ is the total number of legal macro-arms.

\begin{proposition}\label{ens-simple-proposition}
The {\em simple regret} of NS($\epsilon_0$,$\epsilon_l$,$\epsilon_g$) decreases at an exponential rate as $r_T = O(D e^{-2d^2 T p_i})$, where $p_i \geq \epsilon_0 \Pi_{j = 1...n}\frac{\epsilon_l}{K_j} + (1-\epsilon_0)\frac{\epsilon_g}{N}$, $D$ is as in Proposition~\ref{ens-cumulative-proposition}, and $d$ is the minimum difference in reward between an optimal macro-arm and a non-optimal macro-arm 
(proof in Appendix \ref{sec:appendix}).
\end{proposition}

So, in summary, na\"{i}ve sampling has linear cumulative regret (which means that even after a very long number of iterations, it will still pick suboptimal arms with a fixed probability) and exponentially decreasing simple regret (which means that the probability of the arm believed to be the best at the very end of the execution not to be optimal decreases exponentially with the number of iterations executed). This is expected, since, it inherits these properties from $\epsilon$-greedy, which also has linear cumulative regret and exponentially decreasing simple regret (see Appendix \ref{sec:appendix}). 
Also notice that as presented here, $\epsilon$-greedy na\"{i}ve sampling is a strict generalization of $\epsilon$-greedy. If $\epsilon_l = 1.0$ and $\epsilon_g = 0.0$, $\epsilon$-greedy na\"{i}ve sampling is equivalent to an $\epsilon$-greedy policy with parameter $\epsilon_0$. Also, although variations of the $\epsilon$-greedy strategy are known that have logarithmic cumulative regret (e.g., see \citeR{auer2002finite}), we will not explore those in this paper. 

Moreover, notice the interesting inverse relation between simple regret and cumulative regret (already noted by \citeR{bubeck2011pure}). According to the previous propositions, to minimize cumulative regret, we need to make $\epsilon_g$ and $\epsilon_l$ as small as possible, and to minimize simple regret, we need to make $\epsilon_g$ and $\epsilon_l$ as large as possible. So, if we were mostly interested in simple regret in the context of RTS games, this points out that larger values might result in stronger game play. This is echoed in our experimental results, where the best performance was achieved with relatively high values for $\epsilon_l$ (0.4). Notice, moreover, that setting $\epsilon_l$ and $\epsilon_g$ to $1$ (as might seem to be suggested by the results of the propositions), would not work well in practice. The reason is that the proposition results concern a large computational budget (larger than the number of arms). For a smaller computational budget (more realistic in practice), large values of $\epsilon_l$ and $\epsilon_g$ will just result in never sampling any arm more than once, leading to a very poor estimate of their rewards, and thus to low performance. For that reason, our experiments indicate that relatively large (but not all the way to $1$) values of $\epsilon_l$ achieve the best results.

Thus, the upper bounds in the previous propositions are for a sufficiently large number of iterations. However, the key problem in CMABs is that we assume that the number of iterations we can perform is small compared to the number of possible macro-arms. Therefore, it is interesting to analyze the behavior of these strategies when the number of iterations is small. For that purpose, Section \ref{sec:sampling-comparison} presents an empirical comparison of the different strategies presented in this paper.

\subsection{Two-Phase Na\"{i}ve Sampling}

Under the assumption that number of iterations we can perform is much smaller than the total number of macro-arms, $T \ll N$, the global MAB will never reach the point of having all possible macro-arms. Moreover, as pointed out by \citeauthor{shleyfman2014combinatorial}~\citeyear{shleyfman2014combinatorial}, it does not make sense to consider new macro-arms toward the end of the computation budget, since we are not going to have enough time to obtain accurate estimations of their expected reward. This motivates sampling strategies which vary their exploration and exploitation trade-offs over time (e.g., starting with a large probability of exploration, and gradually reducing it). An example of such strategy for regular MABs is the {\em decreasing $\epsilon$-greedy} sampling strategy \cite{auer2002finite}. In this paper, we will focus on the simplest instantiation of these strategies: {\em two-phase} sampling strategies, which instead of gradually changing the probability of exploration, they perform a first ``exploration'' phase to find a set of candidate macro-arms, and in a second phase they try to find the best macro-arm, only amongst those explored during the first phase. An example two-phase strategy for CMABs is LSI \cite{shleyfman2014combinatorial} (described in Section \ref{subsec:lsi}). 

We will write NS($k,\epsilon_0^1,\epsilon_l^1,\epsilon_g^1, \epsilon_0^2, \epsilon_l^2, \epsilon_g^2$) to denote a na\"{i}ve sampling strategy where parameters $\epsilon_0^1, \epsilon_l^1, \epsilon_g^1$ are used during the first $k$ sampling iterations, and $\epsilon_0^2, \epsilon_l^2, \epsilon_g^2$ are used during the rest of the iterations. Following the intuition above, the parameters in the first phase should be geared toward exploring (high values for all the parameters) and the ones in the second phase, should be geared toward exploiting (low values for all the parameters).

If the computation budget $T$ is known ahead of time, $k$ can be set to a fraction $r$ of the total computation budget ($k = rT$). In that case, we will denote the strategy by NS($rT, \epsilon_0^1,\epsilon_l^1,\epsilon_g^1, \epsilon_0^2,\epsilon_l^2,\epsilon_g^2$).

The theoretical analysis of the two-phase na\"{i}ve sampling strategy is very similar to the one-phase case, with the exception of one interesting case (when the first phase lasts for a finite number of iterations $k$, and $\epsilon_0^2 = 0$), which is the only case we will consider here (in all other cases, cumulative regret grows linearly and simple regret decreases exponentially as in the one-phase case). Let us start by bounding the probability of sampling the optimal macro-arm at least once during the first phase.

\begin{proposition}\label{prop:2phase-explore}
In a CMAB with $n$ variables, the probability that after $t$ iterations using a NS$(\epsilon_0, \epsilon_l, \epsilon_g)$ sampling strategy an optimal macro-arm $V^*$ has not been explored at least once, decreases exponentially as a function of $t$, and is at most $(1-p)^{t\epsilon_0}$, where $p = \prod_{i = 1 ... n} \left(\epsilon_l/K_i\right)$ (proof in Appendix \ref{sec:appendix}).
\end{proposition}

This means that in order to maximize the probability of having $V^*$ among the explored macro-arms during the first phase, we want to maximize both $\epsilon_0^1$ and $\epsilon_l^1$. Given Proposition~\ref{prop:2phase-explore}, we can now analyze the behavior when $k$ is a finite number of rounds, and $\epsilon_0^2 = 0$.

\begin{proposition}\label{prop:2phase-cumulative}
The {\em cumulative regret} of NS($k,\epsilon_0^1,\epsilon_l^1,\epsilon_g^1, \epsilon_0^2, 0, \epsilon_g^2$) when $k$ is a constant grows linearly when $T \gg k$: 
$$R_T = O\left(t\left[ (1 -\epsilon_g - q_k + \epsilon_g q_k)d + \epsilon_g D \right]\right)$$
\noindent where $q_k = 1 - (1-p)^{k\epsilon_0^1}$, and $p = \prod_{i = 1 ... n} \left(\epsilon_l^1/K_i\right)$.
(proof in Appendix \ref{sec:appendix}).
\end{proposition}

Notice this is worse than the one-phase case in the limit, since it grows faster.

\begin{proposition}\label{prop:2phase-simple}
The {\em simple regret} of NS($k,\epsilon_0^1,\epsilon_l^1,\epsilon_g^1, \epsilon_0^2, 0, \epsilon_g^2$) when $k$ is a constant and $T \gg k$ is lower bounded by $(1-q_k)d$, where $q_k = 1 - (1-p)^{k\epsilon_0^1}$, and $d$ is difference between the best non-optimal macro-arm and an optimal macro-arm.
(proof in Appendix \ref{sec:appendix}).
\end{proposition}

This means that in the particular case where $\epsilon_l^2 = 0$ and $k$ is a constant, even after a very large number of iterations, the probability of selecting a suboptimal arm at the very end will not approach zero, since it will depend on wether the optimal macro-arm was explored during the first phase or not. 

Moreover, notice that even if in theory, the asymptotic regret bounds seem to be worse than the one-phase strategy, in practice, a two-phase strategy might work better in some scenarios, since what matters in practice (and in the specific application domain of RTS games) is their behavior for small computational budgets (this is evaluated in Section \ref{sec:naive-sampling-comparison}).

\subsection{Na\"{i}ve Sampling Beyond $\epsilon$-greedy}

An interesting question is whether MAB sampling strategies such as UCB1 \cite{auer2002finite}, commonly used in MCTS algorithms can improve over na\"{i}ve sampling using $\epsilon$-greedy. One problem of UCB1 is that it requires exploring each arm at least once (unless strategies like {\em First Play Urgency}, FPU, are used, see \citeR{gelly2006exploration}), which is problematic, since the number of macro-arms is very large. 

Although we do not provide any theoretical results for this strategy, in the experiments below, we experimented with using UCB1 as the sampling strategy for the global MAB in regular na\"{i}ve sampling. We will write NS($\epsilon_0,\epsilon_l,$ UCB1) to denote this strategy. 

\section{Other CMAB Sampling Strategies}\label{sec:other-cmab-strategies}

Two other sampling strategies for CMABs exist in the literature: MLPS \cite{gai2010learning} and LSI \cite{shleyfman2014combinatorial}, which we summarize here. A third algorithm CUCB \cite{chen2013combinatorial} exists, but it is restricted to the specific case where all the variables are boolean (choosing a macro-arm corresponds to choosing a subset of the variables in the CMAB), and thus, we do not include it in our analysis.

\subsection{Matching Learning with Polynomial Storage (MLPS)}

{\em MLPS} (Matching Learning with Polynomial Storage) was presented by \citeauthor{gai2010learning}~\citeyear{gai2010learning} for the problem of multiuser channel allocation. MLPS works in a very similar way to the exploration part of na\"{i}ve sampling. Specifically, MLPS keeps the same $T^t_i(v_i^k)$ and $\overline{\mu}^t_i(v_i^k)$ estimates for the values that each variable can take. The main difference with respect to na\"{i}ve sampling is in the way the macro-arm is selected at each iteration. MLPS assumes that all the variables can take the same values (i.e., $\forall i,j: \mathcal{X}_i = \mathcal{X}_j$), and thus uses the Hungarian algorithm \cite{kuhn1955hungarian} to find the macro-arm that maximizes the expression:
$$W_V^t(n) = \sum_{v_i^k \in V}\overline{\mu}^t_i(v_i^k) + C M \sqrt{\frac{(M+1)\mathit{ln} \, t}{\mathit{min}_{v_i^k \in V} T^t_i(v_i^k)}}$$

Where $M$ is the number of values a variable can take, and $C$ is the exploration parameter (in the original paper, \citeauthor{gai2010learning}~set $C = 1$). In our more general CMAB setting, where each variable has an arbitrary number of values, and where we have an additional function $L$ that determines which macro-arms are legal, the Hungarian algorithm cannot be used directly. Thus, in the experiments presented below, we replaced the Hungarian algorithm with a greedy approach as follows:
\begin{enumerate}
\item Start with an empty macro-arm $V$.
\item Select a random $X_i \in X$ that does not yet have a value.
\item Select the value for $X_i$ which, given the previous selected values, maximizes $W_V^t(n)$.
\item Repeat until all variables have a value.
\end{enumerate}
The previous process is repeated a certain number of times (10 in our experiments), and the iteration that resulted in the highest value is selected. Moreover, we set $M$ in the equation above to be the number of values of the variable that has the most possible values. Although this does not ensure selecting the macro-arm that maximizes $W_V^t(n)$, it is an efficient approach, suitable for real-time games. To distinguish this adapted MLPS strategy from the original MLPS strategy, we will refer to it as MLPS$_{\mathit{greedy}}$.

\subsection{Linear Side Information (LSI)}\label{subsec:lsi}

LSI~\cite{shleyfman2014combinatorial} is a family of two-phase sampling strategies based on the following idea: while na\"{i}ve sampling interleaves exploration and exploitation, LSI splits the computation budget $T = T_g + T_e$ into a first {\em candidate generation} phase (with computation budget $T_g$) and a second {\em candidate evaluation} phase (with computation budget $T_e$). During candidate generation, LSI first collects {\em side information} (analogous to the $\overline{\mu}^t_i(v_i^k)$ estimates in na\"{i}ve sampling), and then, using that information, it generates $k$ candidate macro-arms. During the second phase, LSI uses {\em sequential-halving}~\cite{karnin2013almost} to determine the best of the $k$ macro-arms. During candidate generation, 
the computation budget $T_g$ is divided equally among all the different values of the different variables in the CMAB\@. The different LSI strategies differ in the way these $T_g$ samples are used to collect the side information, and how this side information is used to generate the candidates:
\begin{itemize}
\item LSI$_V$: assuming that there is a value for each variable that is special (in the case of RTS games, the ``no action'' unit-action), the computation budget of each action is used by setting the value of all the other variables to this special value. In this way, LSI obtains an estimate of how much each value of each variable contributes to the global reward (assuming a linear contribution). 
\item LSI$_F$: the computation budget for each value is used by setting random values for all the other variables.
\end{itemize}
Once the candidate generation computation budget is spent, LSI estimates the expected contribution of each value of each variable to the overall reward. Using this estimation, two strategies for generating candidate macro-arms are proposed: 
\begin{itemize}
\item LSI$^e$ (entropy): first, the variables of the CMAB are sorted in decreasing order of entropy, where entropy of a variable is calculated as the entropy of the set of estimated rewards for each of the values of the variable. Intuitively a variable with high entropy is one where the expected reward of its different possible values are very different to each other, while in a variable with low entropy, the expected reward for all of its values will be very similar. Using this order, the variables are then sampled one by one to generate new macro-arms. To sample a value for each variable, the vector of expected rewards for each value of the variable is normalized, so it forms a probability distribution, which is used to generate a value for this variable. Notice that sorting the variables is useful since selecting a value for a variable might prevent selecting certain values for other variables. Thus, sampling first those variables that have a high entropy ensures that the variables that have a larger impact on the expected reward are sampled without having any of their values forbidden by some prior choice.
\item LSI$^u$ (union): instead of sorting the variables, the union of all the values of all the variables that still do not have a value is used for sampling the next value to add to the macro-arm, until the macro-arm is complete (i.e., it has a value for each variable).
\end{itemize}
Once $k$ candidate macro-arms have been generated, sequential-halving is used to determine the best one, with the remaining computational budget. In our experiments, we used the LSI$^e_V$, which has been reported to obtain the best results \cite{shleyfman2014combinatorial}. For a more detailed description of LSI, the reader is referred to the work of \citeauthor{shleyfman2014combinatorial}~\citeyear{shleyfman2014combinatorial}.

\section{Empirical Comparison of CMAB Sampling Strategies}\label{sec:sampling-comparison}

\begin{figure}[t]
    \centering
    \includegraphics[width=0.85\columnwidth]{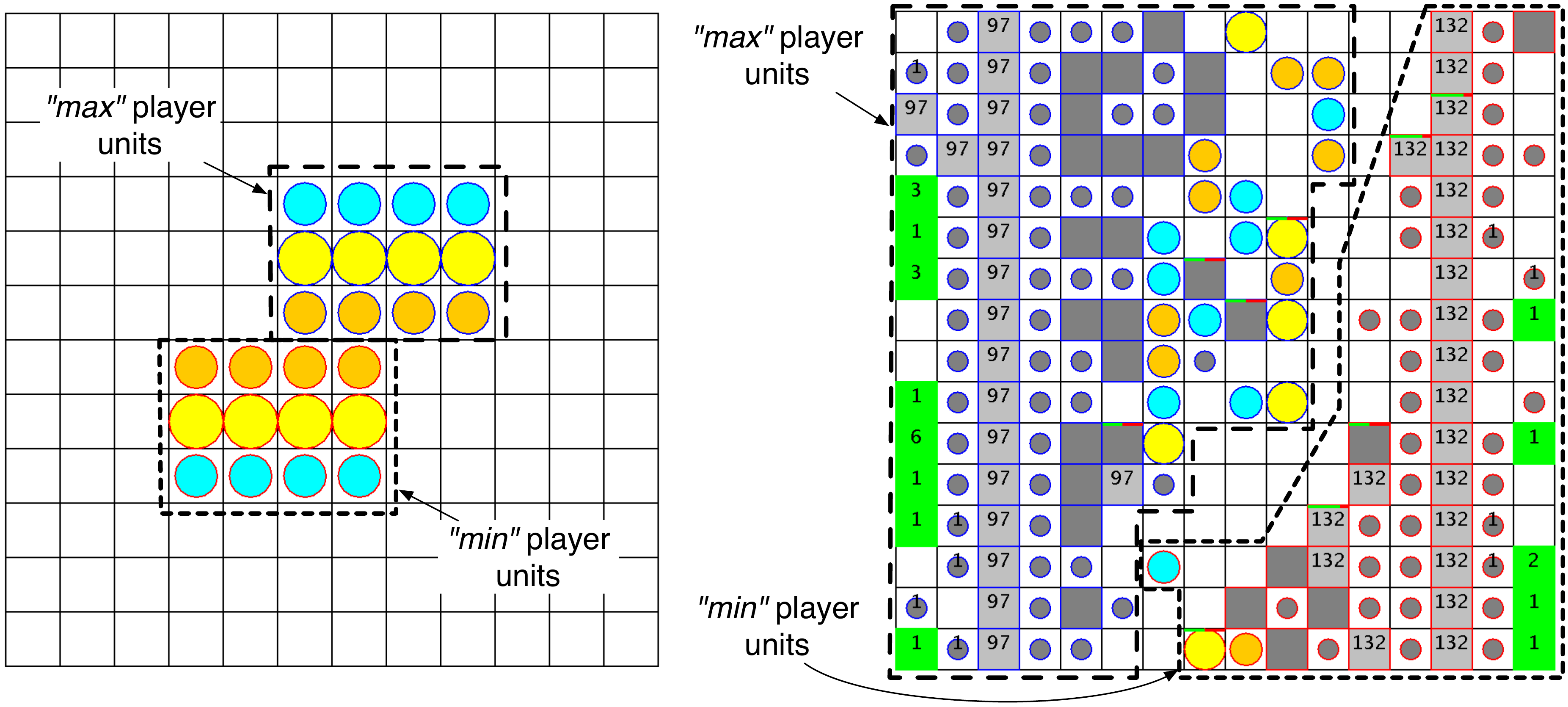}
    \caption{$\mu$RTS situations corresponding to CMAB$_1$ (left) and CMAB$_3$ (right) used in the experiments presented in this paper. Exact situation definitions can be found in the source code link provided in the introduction.}
    \label{fig:cmabs13}
\end{figure}

In order to illustrate the performance of na\"{i}ve sampling compared to other sampling strategies for MABs or CMABs this section presents an empirical comparison. For this comparison, we employed three CMABs (with an increasing number of macro-arms), corresponding to three specific situations in $\mu$RTS:
\begin{itemize}
\item CMAB$_1$: used by \citeauthor{shleyfman2014combinatorial}~\citeyear{shleyfman2014combinatorial} corresponds to the situation shown on the left-hand side of Figure \ref{fig:cmabs13}, from the perspective of the blue player ({\em max}). It has 12 variables, and a total of 10,368 legal macro-arms.
\item CMAB$_2$: which corresponds to the situation depicted in Figure \ref{fig:microRTS} from the perspective of the blue player ({\em max}). CMAB$_2$ has 9 variables (corresponding to the 9 units controlled by the blue player) and a total of 1,008,288 legal macro-arms.
\item CMAB$_3$: corresponds to a larger situation (right-hand side of Figure \ref{fig:cmabs13}, from the perspective of the blue player, {\em max}). This is a $16\times16$ map, with 110 variables (although only 50 can take more than 1 value) and $9.28\times10^{22}$ legal macro-arms\footnote{The number of macro-arms was calculated with the built-in branching factor calculator of $\mu$RTS.}.
\end{itemize}

Notice that these numbers of macro-arms are many orders of magnitude larger than the number of arms typically considered in MABs. We evaluate the following strategies:
\begin{itemize}
\item {\bf $\epsilon$-greedy} (treating the CMABs as if they were actually MABs): we show results for $\epsilon = 0.25$ and $\epsilon = 0.5$ (we tested values between 0.0 and 1.0 at intervals of 0.125 and show the ones that obtained the best results). $V_t^{best}$ is selected based on which arm has been sampled most often at iteration $t$.
\item {\bf UCB1}~\cite{auer2002finite} (also considering MABs): given that the number of macro-arms is larger than the number of samples we can perform, the value of the exploration parameter $C$ of UCB1 has no effect in this experiment (in fact, using UCB1 when there are more arms than the number of iterations that can be run is hopeless, but we included it in our analysis just to set a baseline). However, we set it to $C = 0.05$, which achieved the best results when we combine UCB1 with FPU (below). $V_t^{best}$ is selected based on which arm has been sampled most often at iteration $t$ (ties are resolved by selecting the arm with the highest expected evaluation so far).
\item {\bf UCB1-FPU}~\cite{gelly2006exploration} (also considering MABs): we set the FPU constant to $0.51$, $0.56$ and $0.60$ for each of the three CMABs used in our evaluation. These values achieved the best results in our evaluation (in a deployed system we would not be able to change this value depending on the situation, but we wanted to show the best that UCB1-FPU can achieve in each scenario), and $C = 0.05$. $V_t^{best}$ is selected based on which arm has been sampled most often at iteration $t$.
\item {\bf MLPS$_{\mathit{greedy}}$} (a variation of MLPS, see \citeR{gai2010learning}, as described above): we used $C = 0.005$ for this strategy, which achieved the best results in our experiments.
\item {\bf LSI$_V^e$}~\cite{shleyfman2014combinatorial}: the linear side information strategy described above. We divided the computation budget as $T_g = 0.25 \times T$ and $T_e = 0.75 \times T$, which achieved the best results in our experiments.
\item {\bf NS($\epsilon_0$,$\epsilon_l$,$\epsilon_g$)}: $\epsilon$-greedy na\"{i}ve sampling strategy. We used $\epsilon_0 = 0.8$ (emphasizing exploration), $\epsilon_l = 0.4$, and $\epsilon_g = 0$ (emphasizing the fact that the global MAB is used for exploitation), since they achieved the best results in our experiments. $V_t^{best}$ is selected based on which arm has been sampled most often at iteration $t$.
\end{itemize}

In order to compare the strategies, we evaluate the expected reward of the arm that would be selected as the best at each iteration ($V_t^{best}$), i.e., the simple regret. As the reward function, we use the result of running a Monte Carlo simulation of the game during 100 game cycles (using a random action selection strategy), and then using an evaluation function to the resulting game state.
As the evaluation function, we used one of the built-in function in $\mu$RTS ({\em SimpleSqrtEvaluationFunction3}). Given a state $s$, this evaluation function (inspired by the standard LTD2 function, see \citeR{churchill2012abcd}) assigns a {\em score} to each player ({\em max} and {\em min}) by summing the resource cost of each of her units, weighted by the square root of their health. Then, it produces a normalized evaluation (in the interval $[-1,1]$) as: $E(s) = \frac{2*\mathit{score(max)}}{\mathit{score(min)} + \mathit{score(max)}} - 1$. This is the evaluation function used in all the experiments reported in this paper.

\begin{figure*}[t!]
    \centering
    \includegraphics[width=0.325\textwidth]{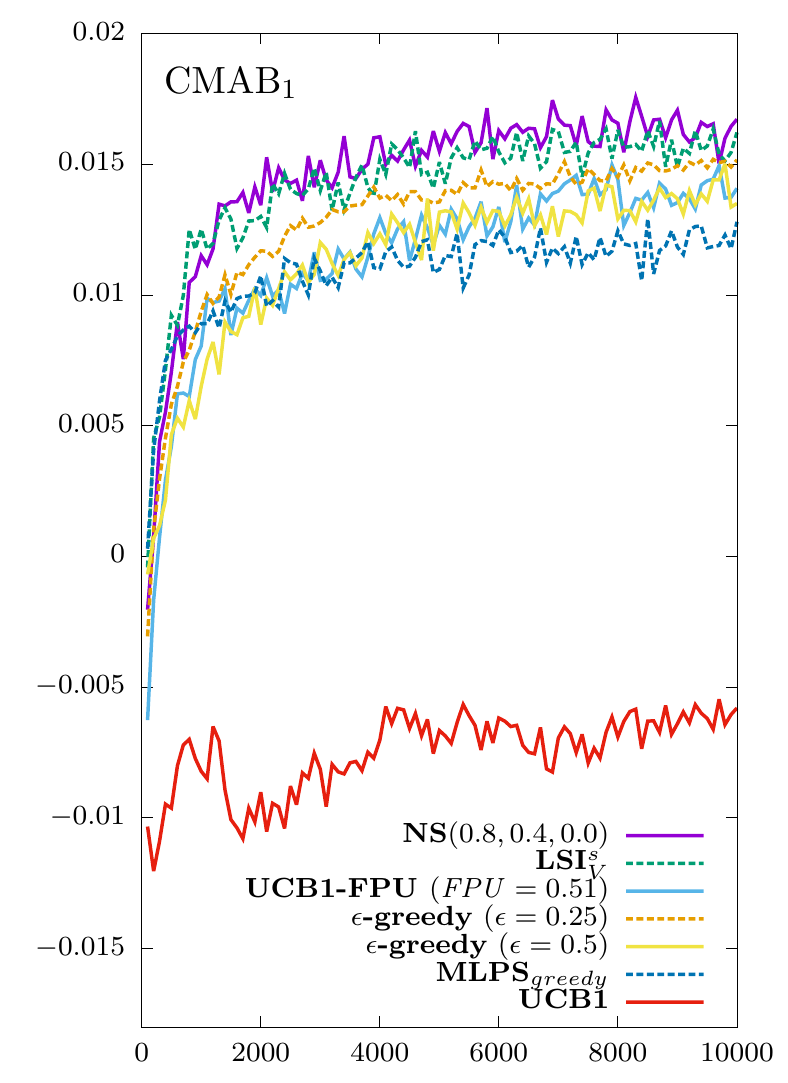}
    \includegraphics[width=0.325\textwidth]{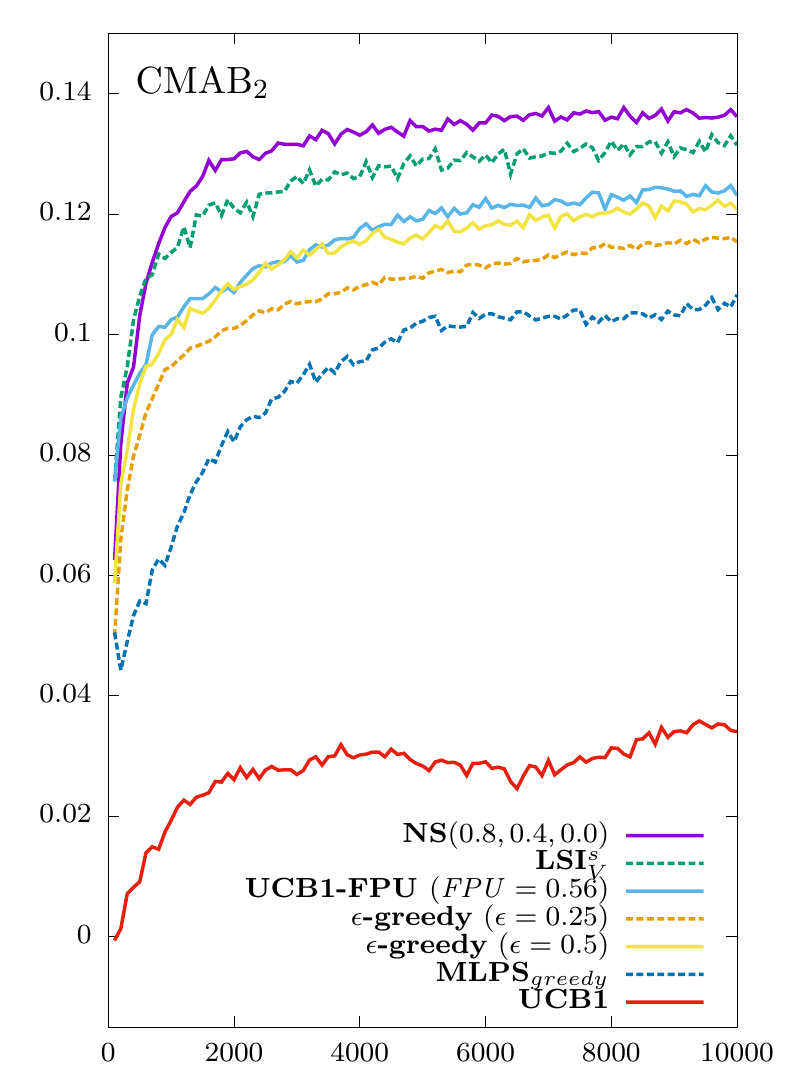}
    \includegraphics[width=0.325\textwidth]{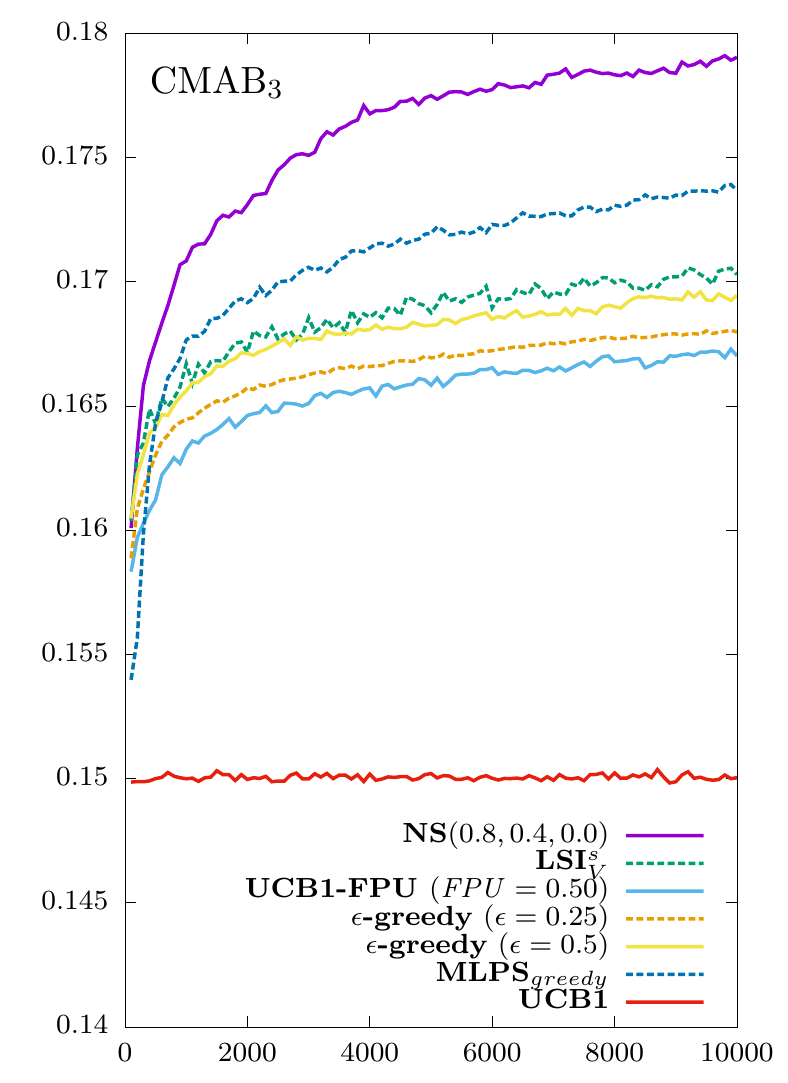}
    \caption{Evolution of the expected reward (vertical axis) of the best arm ($V_t^{best}$) as the computational budget grows (horizontal axis) on three different CMABs. From left to right: CMAB$_1$ has 10,368 legal macro-arms, CMAB$_2$ has 1,008,288 legal macro-arms and CMAB$_3$ has about $9.28\times10^{22}$.}
    \label{fig:cmab1}
\end{figure*}


\begin{table}[t!]
    \centering
    \caption{95\% confidence intervals of the results reported in Figure \ref{fig:cmab1} after 10,000 iterations. We used bold text to highlight the strategies whose intervals overlap with the one achieving the highest reward.}
    \label{tbl:cmab1}
    \small{
    \setlength\tabcolsep{0.15cm}
    \begin{tabular}{|c|c|c|c|} \hline
											& CMAB$_1$ 				& CMAB$_2$ 			& CMAB$_3$ \\ \hline
	\textbf{NS}$(0.8, 0.4, 0.0)$			&  {\bf 0.0151 -  0.0184}		& {\bf 0.1345 - 0.1379}	& {\bf 0.1785 - 0.1796}	   \\ \hline
	LSI$_V^e$								&  {\bf 0.0146 -  0.0178}		& 0.1295 - 0.1334	& 0.1698 - 0.1708	   \\ \hline
	UCB1-FPU								&  {\bf 0.0127 -  0.0155}		& 0.1212 - 0.1250	& 0.1665 - 0.1676	   \\ \hline
	$\epsilon$-greedy ($\epsilon = 0.25$)	&  {\bf 0.0137 -  0.0166}		& 0.1127 - 0.1181	& 0.1665 - 0.1695	   \\ \hline
	$\epsilon$-greedy ($\epsilon = 0.5$)	&  0.0122 -  0.0148		& 0.1185 - 0.1230	& 0.1690 - 0.1699	   \\ \hline
	MLPS$_{\mathit{greedy}}$				&  0.0111 -  0.0144		& 0.1039 - 0.1093	& 0.1731 - 0.1742	   \\ \hline
	UCB1									& -0.0064 - -0.0052		& 0.0278 - 0.0403	& 0.1490 - 0.1511	   \\ \hline
    \end{tabular}
    }
\end{table}

\subsection{Experiment 1: Comparison of CMAB Strategies in CMABs of Increasing Complexity}\label{subsec:experiment1}

Figure \ref{fig:cmab1} shows the average expected reward for a collection of sampling strategies in all three CMABs when sampling between 100 and up to 10,000 iterations (reward ranges from -1 to 1). To measure the performance of the best action generated at each point in time, we compute the average of 200 Monte Carlo simulations of the game during 100 cycles, and then apply the same evaluation function described above. The plots are the average of repeating the experiment 100 times. As can be seen, na\"{i}ve sampling clearly outperforms the other strategies, since the bias introduced by the na\"{i}ve assumption helps in quickly selecting good player-actions. UCB1 basically did a random selection, since it requires exploring each action at least once, and there are more than 10,000 legal macro-arms in all CMABs. UCB1-FPU performed much better, but still significantly below na\"{i}ve sampling. In fact, UCB1-FPU performs similar to a simple $\epsilon$-greedy in CMAB$_1$, better in CMAB$_2$ and significantly worse in CMAB$_3$. MLPS$_{\mathit{greedy}}$ only performed competitively in the larger CMAB$_3$, but still far from na\"{i}ve sampling. LSI$_V^e$ is the strategy that gets closest to na\"{i}ve sampling, being very close in CMAB$_1$; however, in CMAB$_3$, where the number of macro-arms is significantly larger, the advantage of na\"{i}ve sampling is obvious. In order to assess the statistical significance of the results, Table \ref{tbl:cmab1} reports the 95\% confidence intervals of the average expected reward reported in Figure \ref{fig:cmab1} after sampling 10,000 iterations. When the 95\% confidence intervals of two strategies do not overlap, we can say that their difference is statistically significant ($p = 0.05$). As we can see, the difference in reward between na\"{i}ve sampling and the other strategies is statistically significant in CMAB$_2$ and CMAB$_3$, but not in CMAB$_1$. Specifically, these difference become statistically significant in CMAB$_2$ after 1000 iterations, and in CMAB$_3$ after only 300 iterations. This highlights the advantage of na\"{i}ve sampling in larger CMABs.

It is interesting to note that since this evaluation is closely related to {\em simple regret}, rather than {\em cumulative regret}, strategies that perform more exploration tend to work better. That is why, for example, the best performance for $\epsilon$-greedy in the larger CMAB$_2$ and CMAB$_3$ was achieved with a relatively high $\epsilon = 0.5$. Also, this evaluation seems to contradict results reported by \citeauthor{shleyfman2014combinatorial}~\citeyear{shleyfman2014combinatorial}, however, notice that \citeauthor{shleyfman2014combinatorial} used a version of na\"{i}ve sampling that determined the best arm as the one with the highest expected reward so far, rather than selecting the most sampled one so far (as we do here).

Finally, notice that the advantage of na\"{i}ve sampling seems to increase in larger CMABs (e.g., CMAB$_3$). This is because na\"{i}ve sampling exploits the structure in the domain, and if a value for a given variable is found to obtain a high reward in average, then other macro-arms that contain such value are likely to be sampled. Thus, it exploits the fact that macro-arms with similar values might have similar expected rewards. MLPS also exploits this fact, but since it does not keep a separate {\em global MAB}, as na\"{i}ve sampling does, it cannot pinpoint which was the exact combination of values that achieved the highest expected reward.

We would like to note that there are existing strategies, such as HOO~\cite{bubeck2008online}, designed for continuous actions, that can exploit the structure of the action space, as long as it can be formulated as a topological space. Attempting such formulation, and comparing with HOO is part of our future work. 

\subsection{Experiment 2: Variations of Na\"{i}ve Sampling}\label{sec:naive-sampling-comparison}

In this section we compare a variety of na\"{i}ve sampling configurations using the same methodology used in the previous subsection. We used the following configurations (in all of them $V_t^{best}$ is selected based on which arm has been sampled most often at iteration $t$):

\begin{itemize}
\item {\bf NS($\epsilon_0$,$\epsilon_l$,$\epsilon_g$)}: we employed the same values as before ($\epsilon_0 = 0.8$, $\epsilon_l = 0.4$, and $\epsilon_g = 0.0$).
\item {\bf NS($rT, \epsilon_0^1,\epsilon_l^1,\epsilon_g^1, \epsilon_0^2,\epsilon_l^2,\epsilon_g^2$)}: after exploring the space of value combinations at intervals of $0.1$, the values that performed better are:
	\begin{itemize}
	\item $r = 0.6$, $\epsilon_0^1 = 0.8$, $\epsilon_l^1 = 0.4$, $\epsilon_g^1 = 0.0$, and $\epsilon_0^2 = 0.0$, $\epsilon_l^2 = 0.0$, $\epsilon_g^2 = 0.0$. This means doing standard na\"{i}ve sampling for 60\% of the computation budget, and then do pure exploitation during the remaining 40\%.
	\item $r = 0.6$, $\epsilon_0^1 = 0.8$, $\epsilon_l^1 = 0.4$, $\epsilon_g^1 = 0.0$, and $\epsilon_0^2 = 0.0$, $\epsilon_l^2 = 0.0$, $\epsilon_g^2 = 0.2$. This means doing standard na\"{i}ve sampling for 60\% of the computation budget, and then do $\epsilon$-greedy with $\epsilon = 0.2$ over the set of macro-arms explored during the first phase.
	\end{itemize}
\item {\bf NS($\epsilon_0,\epsilon_l,$ UCB1)}: we used $\epsilon_0 = 0.8$, $\epsilon_l = 0.4$, and UCB1 (with $C = 0.005$) as the strategy to sample the global MAB\@.
\end{itemize}

\begin{figure*}[t!]
    \centering
    \includegraphics[width=0.325\textwidth]{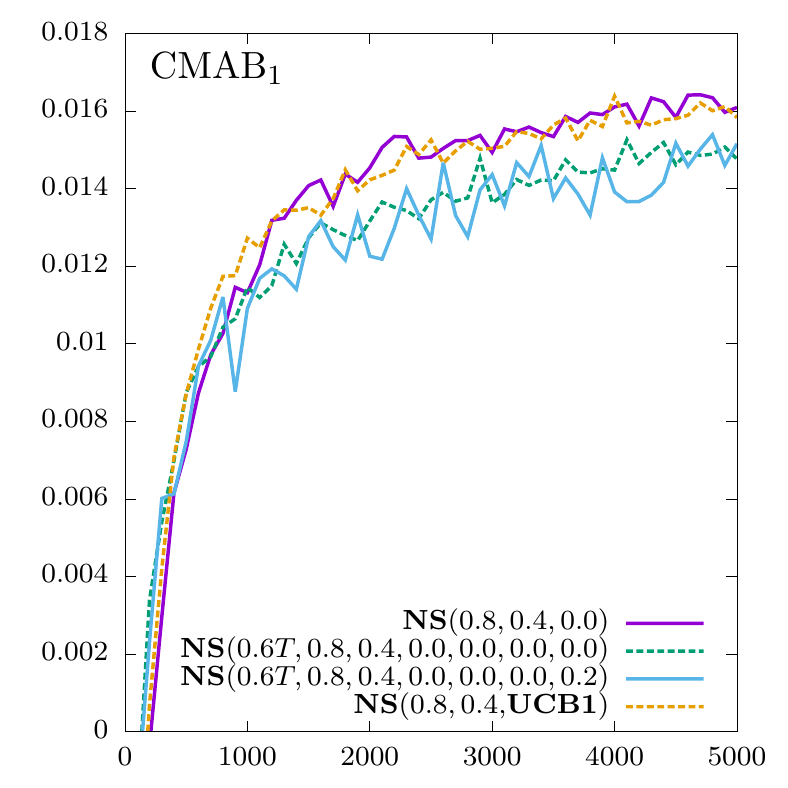}
    \includegraphics[width=0.325\textwidth]{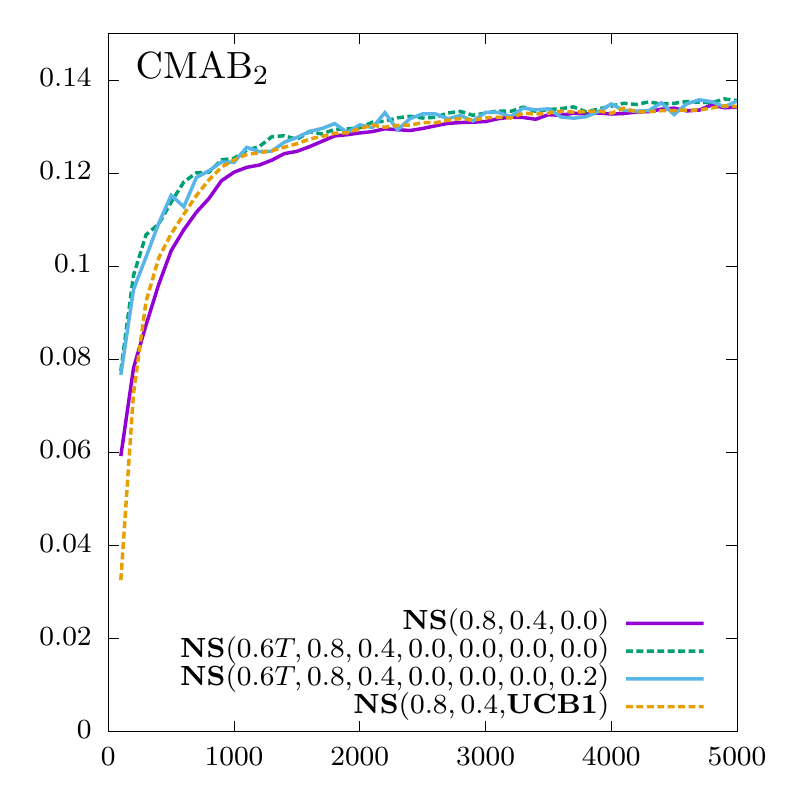}
    \includegraphics[width=0.325\textwidth]{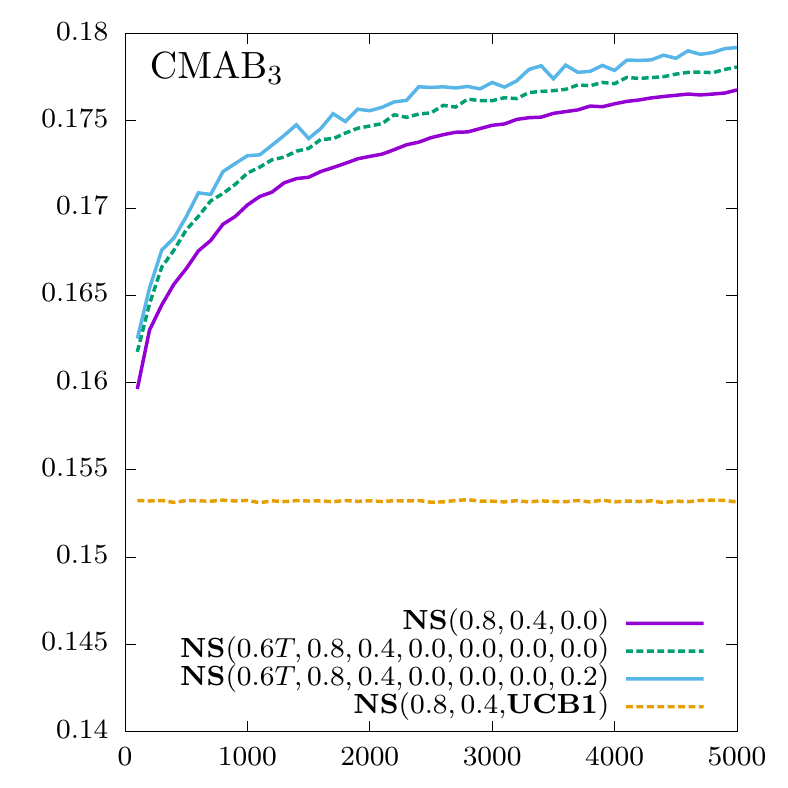}
    \caption{Evolution of the expected reward (vertical axis) of the best arm ($V_t^{best}$) as the computational budget grows (horizontal axis) on the same three CMABs as Figure \ref{fig:cmab1}, for three variants of na\"{i}ve sampling.}
    \label{fig:cmab1-naive}
\end{figure*}


\begin{table}[t!]
    \centering
    \caption{95\% confidence intervals of the results reported in Figure \ref{fig:cmab1-naive} after 5,000 iterations. Statistically significant differences are only observed in CMAB$_3$, highlighted in bold.}
    \label{tbl:cmab1-naive}
    \small{
    \setlength\tabcolsep{0.15cm}
    \begin{tabular}{|c|c|c|c|} \hline
													& CMAB$_1$ 				& CMAB$_2$ 			& CMAB$_3$ \\ \hline
	\textbf{NS}$(0.8,0.4,0.0)$		 				& 0.0149 - 0.0173		& 0.1326 - 0.1357	& 0.1760 - 0.1775	\\ \hline
	\textbf{NS}$(0.6T, 0.8,0.4,0.0, 0.0,0.0,0.0)$	& 0.0136 - 0.0159		& 0.1338 - 0.1374	& 0.1777 - 0,1785 	\\ \hline
	\textbf{NS}$(0.6T, 0.8,0.4,0.0, 0.0,0.0,0.2)$	& 0.0140 - 0.0163		& 0.1337 - 0.1373	& {\bf 0.1788 - 0.1796}	\\ \hline
	\textbf{NS}$(0.8,0.4,$UCB1$)$					& 0.0147 - 0.0170		& 0.1324 - 0.1363	& 0.1520 - 0.1543	\\ \hline
    \end{tabular}
    }
\end{table}

Results are shown in Figure \ref{fig:cmab1-naive} in all three CMABs when sampling between 100 and up to 5,000 iterations, evaluated as in the previous experiment (reward also ranges from -1 to 1). Results show that the relative performance of na\"{i}ve sampling variants depends on the specific CMAB\@. For example, in CMAB$_1$, with a ``small'' branching factor (10,368), a two-phase approach seems to perform worse than standard na\"{i}ve sampling, or UCB-based na\"{i}ve sampling (which perform almost identically). In CMAB$_2$, which has a larger branching factor, the two-phase approach starts to pay off, and performs just slightly better than the other two approaches. Finally, in CMAB$_3$, with a very large branching factor (about $9.28 \times 10^{22}$), the two-phase approach clearly outperforms the other two variants (with the \textbf{NS}$(0.6T, 0.8,0.4,0.0, 0.0,0.0,0.2)$ outperforming all other approaches). Moreover, it is interesting to note that in CMAB$_3$, using a UCB1 sampling strategy for the global MAB does not work, since the branching factor is so large, that the local MABs never select the same macro-arm twice, and thus, the number of arms in the global MAB is larger than the number of times UCB1 is called. However, notice that for very small computation budgets (smaller than 1000), the two-phase approach outperforms standard na\"{i}ve sampling in all CMABs (we will get back to this point later in Section \ref{sec:experiment5}). Table \ref{tbl:cmab1-naive} reports the 95\% confidence intervals of the values obtained by these different strategies after 5,000 iterations, showing that differences are only statistically significant in CMAB$_3$, where a two phase strategy dominates all the others. In fact in CMAB$_3$, the difference in reward of both two phase strategies with respect to standard na\"{i}ve sampling is statistically significant as early as after 100 iterations. In CMAB$_2$ the different is initially statistically significant, but it stops being so after 600 iterations.

The conclusion is that for very large branching factors, a two-phase approach pays off, since the last iterations are spent just trying to narrow down, from the set of macro-arms already sampled, which are the best. When the number of macro-arms is not that large, this does not appear to yield any benefit with respect to standard na\"{i}ve sampling.

\section{Monte Carlo Search based on CMABs for RTS Games}\label{sec:mcts}

As mentioned in Section \ref{sec:cmab}, although many authors have argued for the need of Monte Carlo Tree Search algorithms that use MAB strategies that minimize simple regret instead of cumulative regret, in this paper we will use two standard Monte Carlo search algorithms to evaluate the different CMAB strategies in the context of RTS games: 
\begin{itemize}
\item a Monte Carlo Tree Search (MCTS) approach (since RTS games involve simultaneous and durative actions, we used the MCTS approach described in our previous work, see~\citeR{ontanon2013combinatorial}), described below, and 
\item a plain Monte Carlo (MC) search approach (which was implemented basically by limiting the depth of the tree in MCTS to 1).
\end{itemize}

The main difference between MC and MCTS in our context is that the MC approach does not construct a game tree, as the MCTS approach does. Moreover, some strategies, such as LSI, require knowing the sampling budget before-hand, and thus cannot be used in the context of MCTS (since we cannot anticipate the budget for any tree node except for the root). 
%
%
We used Na\"{i}veMCTS \cite{ontanon2013combinatorial} as our MCTS approach, specifically designed for RTS games. 

The first consideration that Na\"{i}veMCTS does is that unit-actions in an RTS game are durative (they might take several game cycles to complete). For example, in $\mu$RTS, a {\em worker} takes 10 cycles to move one square in any of the 4 directions, and 200 cycles to build a {\em barracks}. This means that if a player issues a move action to a {\em worker}, no action can be issued to that {\em worker} for another 10 cycles. Thus, there might be cycles in which one or both players cannot issue any actions, since all the units are busy executing previously issued actions. The game tree generated by Na\"{i}veMCTS takes this into account, using the same idea as the ABCD ($\alpha$-$\beta$ Considering Durations) algorithm \cite{churchill2012abcd}.

Na\"{i}veMCTS is designed for deterministic two-player zero sum games, where one player, $max$, attempts to maximize the evaluation function $\rho$, and the other player, $min$, attempts to minimize it. Na\"{i}veMCTS differs from other MCTS algorithms in the way nodes are selected and expanded in the tree (the {\em SelectAndExpandNode} procedure in Section \ref{subsec:mcts}).

\begin{algorithm}[tb] \caption{SelectAndExpandNode($n_0$)}\label{alg:naive} 
\begin{algorithmic}[1]
\medskip
\IF {$\text{canMove}(max, n_0.state)$}
\STATE $player = max$
\ELSE
\STATE $player = min$
\ENDIF
\STATE $\alpha = \text{Na\"{i}veSampling}(n_0.state, player)$
\IF {$\alpha \in n_0.children$}
\RETURN $\text{Na\"{i}veSelectAndExpandNode}(n_0.child(\alpha))$
\ELSE
\STATE $n_1 = \text{newTreeNode}(\text{fastForward}(n_0.state, \alpha))$
\STATE $n_0.\text{addChild}(n_1, \alpha)$
\RETURN $n_1$
\ENDIF
\end{algorithmic}
\end{algorithm}

The {\em SelectAndExpandNode} process for Na\"{i}veMCTS is shown in Algorithm \ref{alg:naive}. The process receives a game tree node $n_0$ as the input parameter, and lines 1-5 determine whether this node $n_0$ is a {\em min} or a {\em max} node (i.e. whether the children of this node correspond to moves of player $min$ or of player $max$). Then, line 6 uses na\"{i}ve sampling to select one of the possible player-actions of the selected player in the current state. If the selected player-action corresponds to a node already in the tree (line 8), then {\em SelectAndExpandNode} is recursively applied from that node (i.e. the algorithm goes down the tree). Otherwise (lines 10-12), a new node is created by executing the effect of player-action $\alpha$ in the current game state using the {\em fastForward} function. {\em fastForward} simulates the evolution of the game until reaching a decision point (when any of the two players can issue an action, or until a terminal state has been reached). This new node is then returned as the node from where to perform the next simulation.


A final consideration is that RTS games are simultaneous-action domains, where more than one player can issue actions at the same instant of time. Algorithms like minimax might result in under or overestimating the value of positions, and several solutions have been proposed \cite{KovarskyB05heuristic,SaffidineFinnssonBuro2012AAAI}. However, we noticed that this had a very small effect on the practical performance of our algorithm in RTS games, so we have not incorporated any of these techniques into Na\"{i}veMCTS.

\section{Experimental Results in the Context of Game Tree Search}\label{sec:experiments}

The following subsections present three separate experiments aimed at evaluating different CMAB sampling strategies in the context of Game Tree search.

\subsection{Experimental Setup}

In order to evaluate the performance of the different CMAB sampling strategies, as before, we employed $\mu$RTS. 
$\mu$RTS games used in this paper are fully observable and deterministic, but still capture several defining features of full-fledged RTS video games: durative and simultaneous actions, large branching factors, resource allocation, and real-time combat.

%

\begin{figure}[t!]
    \centering
    \includegraphics[width=0.9\columnwidth]{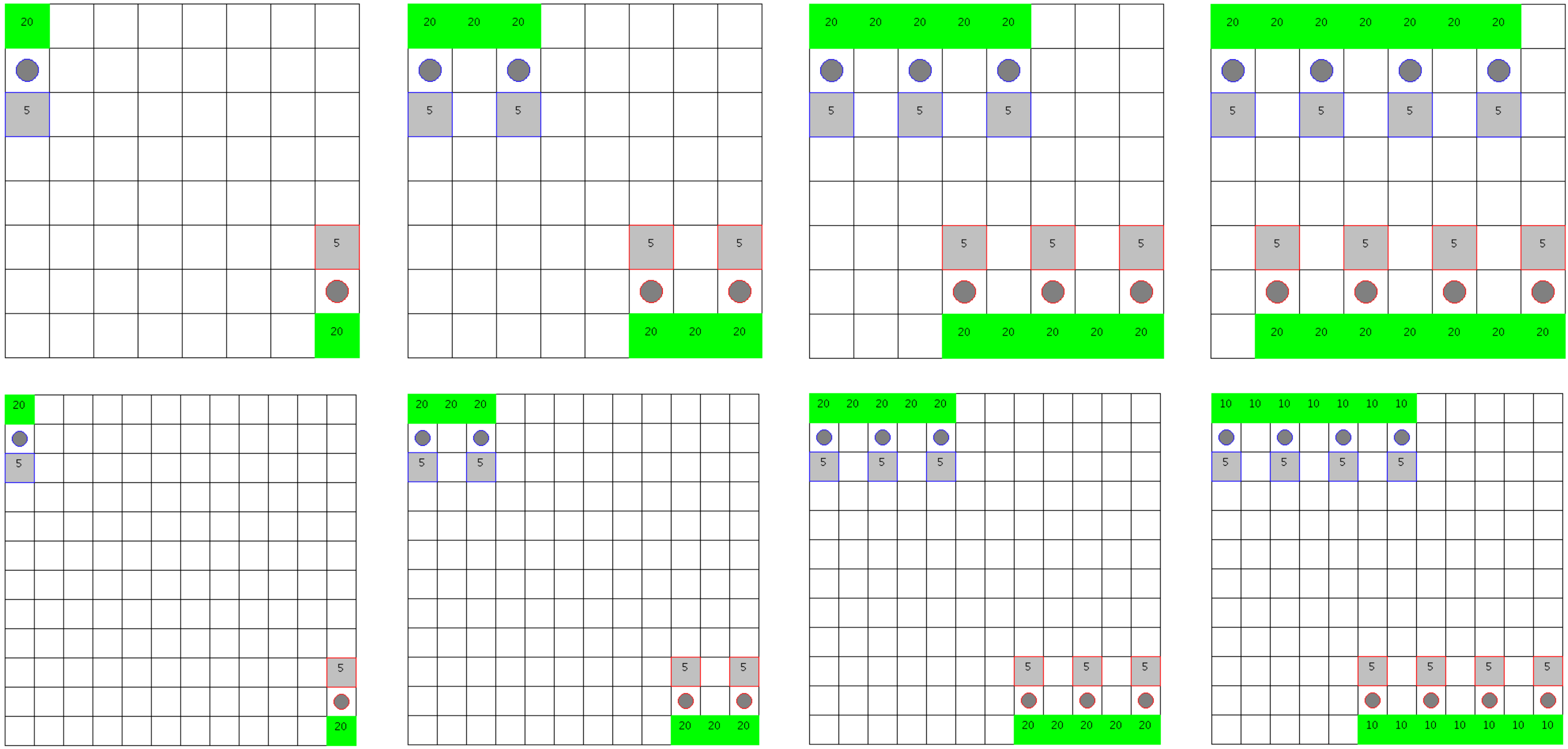}
    \caption{The eight maps used in our experimental evaluation, the top four maps are $8 \times 8$ cells in size, while the bottom four are $12 \times 12$. For each size, we employed maps with a varying number of starting bases and workers, resulting in games with different average branching factors, and average lengths. {\em max} always starts at the top, and {\em min} at the bottom; each player start with 5 resources (the number displayed on each player base); and each resource mine has 20 resources available.}
    \label{fig:maps}
\end{figure}

We employed eight different $\mu$RTS maps, that result in games of different average branching factors (Figure \ref{fig:maps}):
\begin{itemize}
\item Four 8x8 maps ({\bf 8x8-1base}, {\bf 8x8-2base}, {\bf 8x8-3base}, and {\bf 8x8-4base}), shown in the top half of Figure \ref{fig:maps}. In the simplest of them ({\bf 8x8-1base}), each player starts with one base and one worker, and near a single resource mine. In the most complex of them ({\bf 8x8-4base}), each player starts with four bases and four workers, right next to a row of 7 resource mines. 
\item Four 12x12 maps ({\bf 12x12-1base}, {\bf 12x12-2base}, {\bf 12x12-3base}, and {\bf 12x12-4base}), shown in the bottom half of Figure \ref{fig:maps}, analogous to the 8x8 maps, but where players start further apart (given the larger dimensions), thus increasing the average length of a game and allowing a larger maximum number of units in the map.
\end{itemize}

%

We performed two experiments to assess the performance of each of the CMAB sampling strategies described above when used in MC and MCTS in the context of RTS games:

\begin{itemize}
\item {\bf Branching Factor Analysis}: to measure the complexity of the games in each of the 8 maps (and thus compare the results obtained here to those reported in {\em Experiment 1}, see Section \ref{subsec:experiment1}), average game length and branching factor in each map were analyzed.


\item {\bf Round-Robin Analysis}: we selected some of the top performing configurations from the previous experiment, and we ran a round-robin tournament where each configuration played against all others in all the different maps.
\end{itemize}

Games were limited to 3000 cycles, after which the game was considered a draw. Moreover, in the MC and MCTS implementations in $\mu$RTS, when the computation budget is set to $T$ playouts, this does not mean that each decision is made by running MC or MCTS for $T$ playouts. Instead, what this means is that each bot has a computation budget of $T$ playouts per game cycle. Thus, in situations where a bot does not need to issue an action during a few cycles in a row (e.g., because all of its units are already busy), a bot can launch an execution of MC or MCTS that spreads over several game cycles (i.e., the bot starts a search process in the first game cycle, and continues the search during the subsequent game cycles until it needs to produce an action). The following subsections describe the results of each of the experiments.

\subsection{Experiment 3: Branching Factor Analysis}

\begin{table*}[t!]
    \centering
    \caption{Median, Average and Maximum branching factor encountered in each of the eight maps during Experiment 2. $^*$ some branching factor calculations timed out, median/average/max taken of the ones that did not time out.}
    \label{tbl:branching}
    \small{
    \begin{tabular}{|c|ccc|cc|} \hline
					& \multicolumn{3}{c|}{{\em Branching}} 					& \multicolumn{2}{c|}{{\em Game Length}} \\ \hline
	{\em Map} 		& {\em Median} & {\em Average} & {\em Max}				& {\em Average Cycles} & {\em Average Decision Cycles} \\ \hline
	8x8-1base 		& $14.50$ & $1466.35$ & $2.65\times10^{5}$				& $428.33$ & $48.93$ \\ \hline
	8x8-2base 		& $84.00$ & $1.87\times10^{6}$ & $2.49\times10^{9}$		& $474.33$ & $58.93$ \\ \hline
	8x8-3base 		& $106.00$ & $4.53\times10^{5}$ & $1.75\times10^{8}$	& $380.33$ & $52.10$ \\ \hline
	8x8-4base 		& $342.50$ & $9.52\times10^{5}$ & $3.74\times10^{8}$	& $267.00$ & $38.60$ \\ \hline
	12x12-1base 	& $30.50$ & $1.23\times10^{6}$ & $6.14\times10^{8}$		& $972.67$ & $158.17$ \\ \hline
	12x12-2base 	& $112.00$ & $1.56\times10^{12}$ & $2.90\times10^{15}$	& $893.27$ & $151.00$ \\ \hline
	12x12-3base$^*$ 	& $546.75$ & $1.60\times10^{15}$ & $4.98\times10^{18}$	& $804.00$ & $135.3$ \\ \hline
	12x12-4base$^*$ 	& $15330.00$ & $1.08\times10^{17}$ & $8.07\times10^{19}$	& $777.80$ & $126.13$ \\ \hline
    \end{tabular}
    }
\end{table*}

\begin{figure}[t!]
    \centering
    \includegraphics[width=0.47\columnwidth]{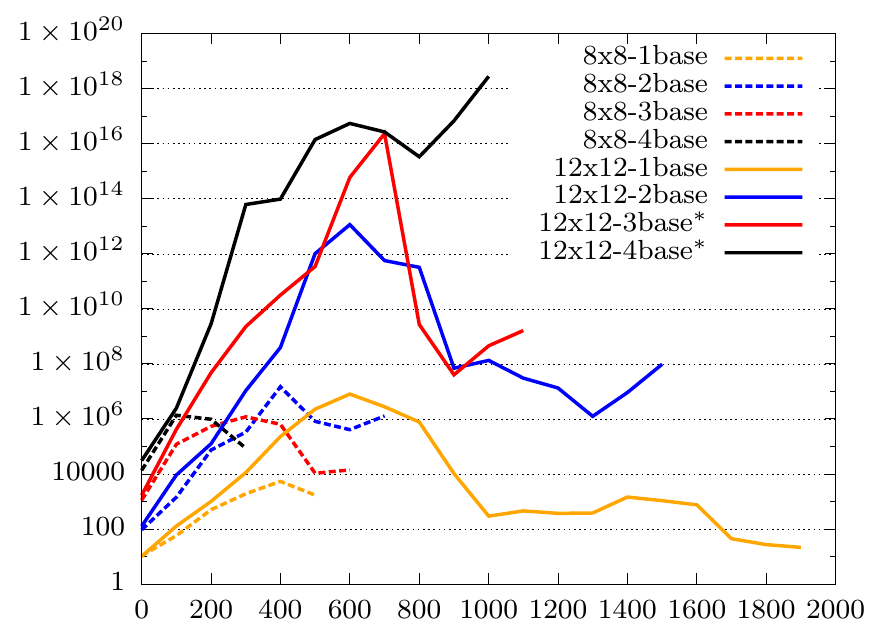}
    \includegraphics[width=0.45\columnwidth]{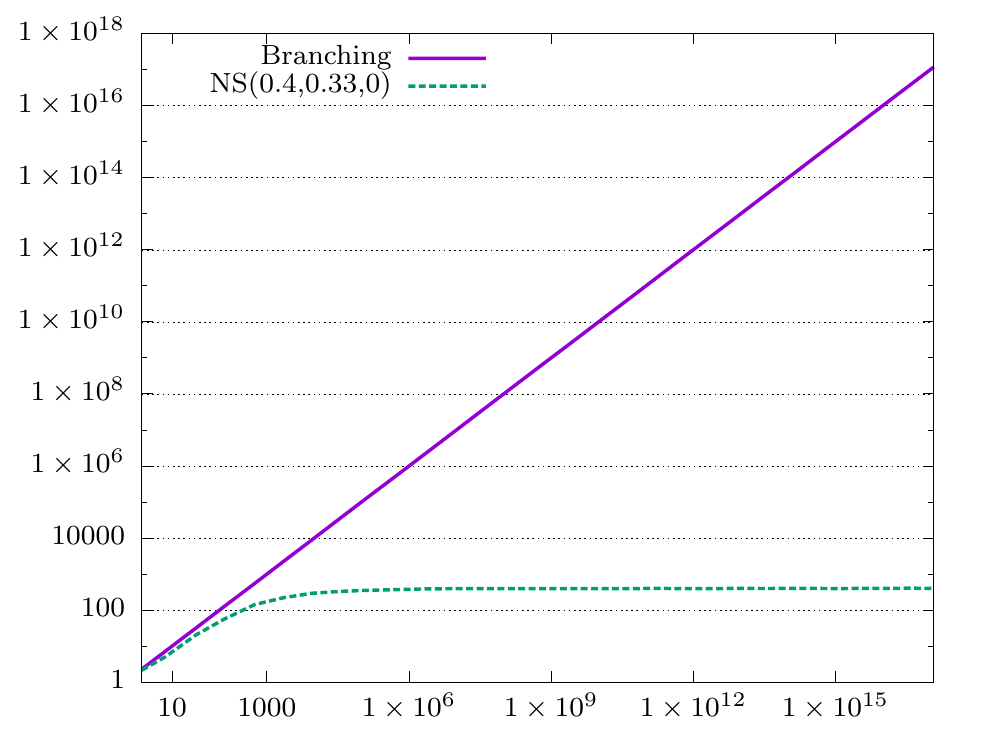}
    \caption{{\bf Left}: average branching factor over the course of games in each of the 8 maps; The horizontal axis represents time (in game cycles), and the vertical axis is the average branching factor. {\bf Right}: comparison of the number of macro-arms explored by {\bf NS}($0.4$,$0.33$,$0$) with a budget of 1000 playouts as the branching factor grows. $^*$ some branching factor calculations timed out, average taken of the ones that did not time out.}
    \label{fig:branching}
\end{figure}

We recorded the branching factor from the point of view of both players by making an AI that uses MC search (with a computation budget of 1000 playouts per frame and {\bf NS}($0.4$,$0.33$,$0$) as the CMAB strategy) play five games against each of three of the scripted AIs that come with $\mu$RTS ({\bf RandomBiased}, {\bf LightRush} and {\bf WorkerRush}) in each one of the 8 maps (a total of $3 \times 8 \times 5 = 120$ games). The scripted AI always played as player {\em max}, and the MC AI always played as player {\em min}. We selected these different scripted AIs, just to have a variety of games, in order to gather a variety of game states to estimate the branching factor from.

Table \ref{tbl:branching} shows the median, average, and maximum branching factors encountered during this experiment for each of the eight maps. 
The right-hand side of Table \ref{tbl:branching} shows the average game length in terms of both game cycles, and ``Decision Cycles'' (where a ``Decision Cycle'' is a game cycle where there was at least one idle unit for which a player had to produce an action). 
We can see that in the simplest map ({\bf 8x8-1base}, used in the past for evaluation of different CMAB strategies, see \citeR{ontanon2013combinatorial} and \citeR{shleyfman2014combinatorial}), branching factors do not grow very large (the average is 1466.35, with a median of 14.50). However, as we increase the number of bases, and especially if we use the larger 12x12 maps, the branching factor grows very rapidly. In the extreme cases of the {\bf 12x12-3base} and {\bf 12x12-4base} maps, some of the branching factors were too large to be computed in a reasonable amount of time. For example, in the {\bf 12x12-4base} map, the average branching factor for the states where we could actually compute it (we set a timeout of 2 hours of CPU time to calculate the branching factor of a game state) was $1.08\times10^{17}$, and median 15330.00 (which means that half of the times the branching factor was larger than 15330.00). In 12x12 maps, the branching factor is smaller than 1000 66.55\% of the times, it is between 1000 and one million 19.06\% of the times, and it is larger than one million 14.39\% of the times. 
The left-hand side of Figure \ref{fig:branching} shows the average branching factor over time for each of the 8 maps. We can see that branching factor starts small at the beginning, when there are few units in the map, and then grows very rapidly. Branching factor tends to decrease toward the end game, since players destroy each other's units during the game. 
Moreover, the right-hand side of Figure \ref{fig:branching} shows the number of macro-arms explored by {\bf NS}($0.4$,$0.33$,$0$) with a budget of 1000 playouts as the branching factor grows. As can be seen, {\bf NS}($0.4$,$0.33$,$0$) flattens out at exploring about 400 macro-arms (corresponding to $\epsilon_0 = 0.4$). In average, {\bf NS}($0.4$,$0.33$,$0$) explored 36.43\% of all the possible macro-arms (notice that even if the percentage of explored macro-arms is nearly 0 for those game states with large branching factor, 66.55\% of those have branching factor smaller than 1000). 

Finally, in the Java $\mu$RTS implementation of these algorithms, running MC search takes about 271ms to run 1000 playouts in 8x8 maps, and about 615ms in 12x12 maps on an Intel Core i7 3.1GHz (the sampling strategy makes little difference, since the main bottleneck is running the forward model to run the playouts). In an optimized C++ implementation, we thus estimate that between 200 and 1000 playouts would be feasible to be run on real time (10 to 24 frames per second) in a commercial RTS game depending on the complexity of the game state.

\subsection{Experiment 4: Round-Robin Analysis}

\begin{figure*}[t!]
    \centering
    \includegraphics[width=0.325\textwidth]{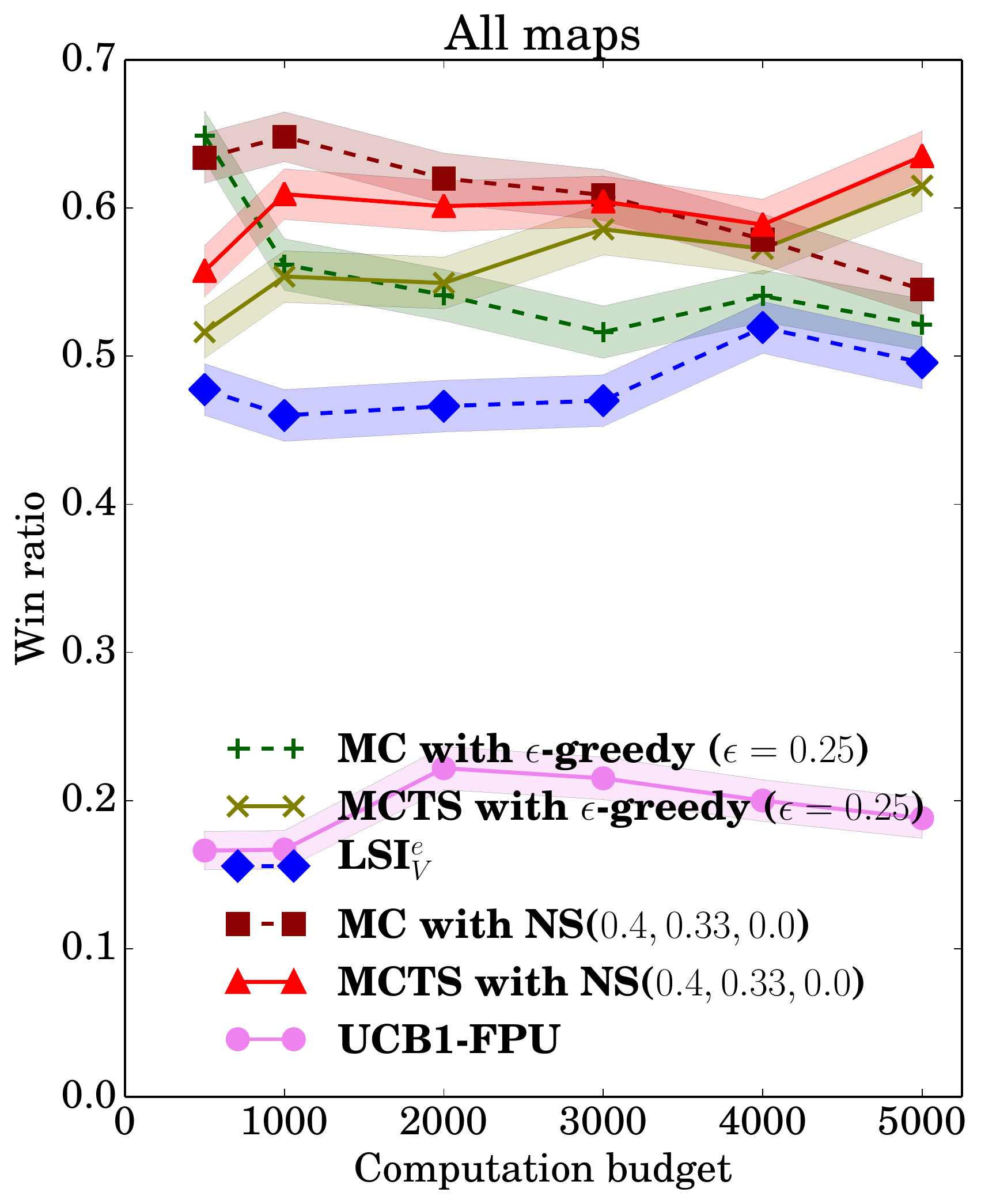}
    \includegraphics[width=0.325\textwidth]{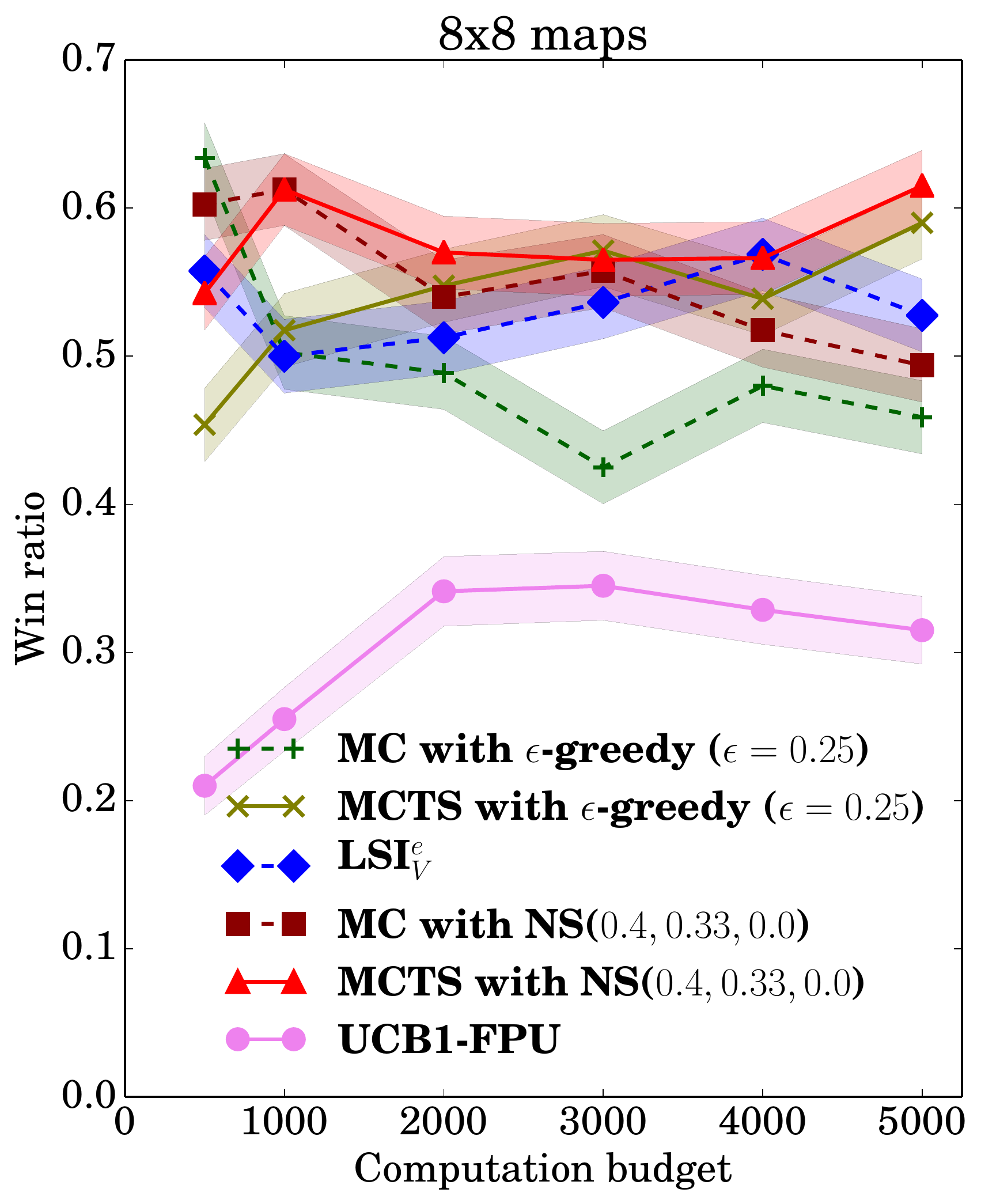}
    \includegraphics[width=0.325\textwidth]{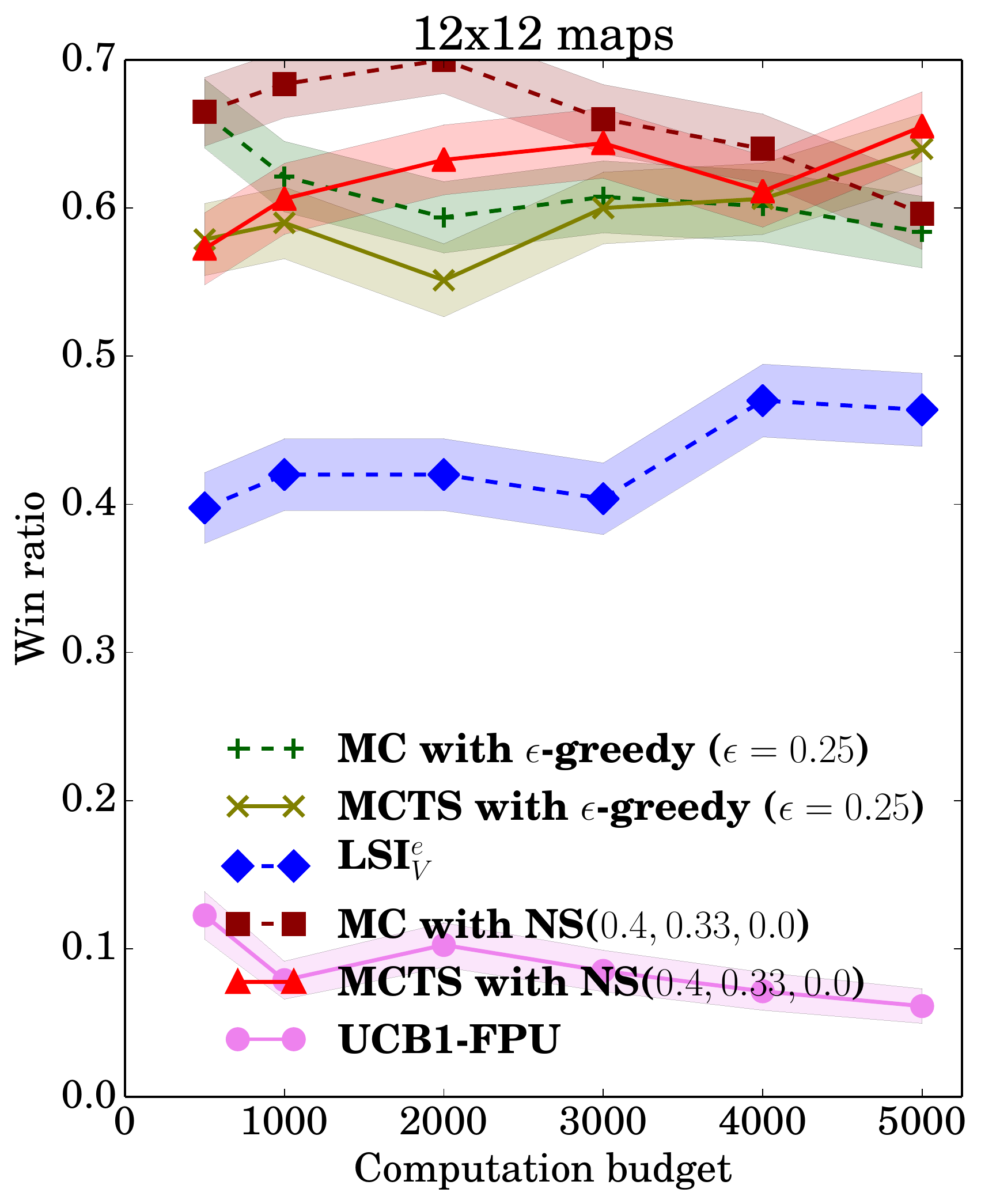}
    \caption{Win ratio of different approaches when doing a round-robin tournament with different computation budgets: averaged over all maps (left), only the 8x8 maps (center) and only the 12x12 maps (right).}
    \label{fig:round-robin}
\end{figure*}


We performed a series of round-robin tournaments involving six different AIs:
\begin{itemize}
\item {\bf MC with $\epsilon$-greedy} ($\epsilon = 0.25$): Monte Carlo search using $\epsilon$-greedy as the tree policy.
\item {\bf MCTS with $\epsilon$-greedy} ($\epsilon = 0.25$): MCTS using $\epsilon$-greedy as the tree policy.
\item {\bf MC with LSI$_V^e$}~\cite{shleyfman2014combinatorial}: plain Monte Carlo search using LSI$_V^e$. Notice that LSI cannot be used with MCTS, since it needs to know in advance the computation budget to be used in a given node of the tree.
\item {\bf MC with NS$(0.4, 0.33, 0.0)$}: Monte Carlo search using na\"{i}ve sampling, NS$(0.4, 0.33,$ $0.0)$, as the sampling policy.
\item {\bf MCTS with NS$(0.4, 0.33, 0.0)$}: Monte Carlo Tree Search using NS$(0.4, 0.33, 0.0)$ as the tree policy.
\item {\bf UCB1-FPU}: Monte Carlo Tress Search using a UCB1 sampling strategy with an FPU constant set to: $\mathit{FPU} = f \times \rho(s) + (1-f)$, where $\rho(s)$ is the value of the evaluation function applied to the game state in the current game state, and $f = 0.95$, set empirically (intuitively, this basically sets the FPU constant to slightly higher than the value of the current game state, so that if actions are found early that improve the evaluation function, those are explored right away).
\end{itemize}

Notice that we test both $\epsilon$-greedy and na\"{i}ve sampling using an MC and a MCTS search algorithm, in order to separate the performance that comes from the sampling strategy from the performance that comes from the search algorithm.

In each round-robin tournament, each AI played 20 games against each other AI (10 times as player {\em max}, and 10 times as player {\em min}) in each of the 8 maps, resulting in a total of 2400 games per tournament. We played six such round-robin tournaments, using a computation budget of 500, 1000, 2000, 3000, 4000 and 5000 playouts per game cycle respectively.

The left hand side of Figure \ref{fig:round-robin} shows the win ratio (vertical axis) of each of the AIs as a function of the computational budget used in each of the tournaments (horizontal axis). ``Win ratio'' was calculated as the average score that bots got in each game, scoring $1$ for winning, $0.5$ for drawing, and $0$ for losing the game. The bands around the plots represent the 95\% confidence interval in the win ratio. Thus, when the bands of two plots do not overlap, their difference is statistically significant with $p = 0.05$. As can be seen, {\bf MC with $\epsilon$-greedy} won the 500 playouts tournament (but with a difference that is not statistically significant), {\bf MC with NS$(0.4, 0.33, 0.0)$} won the 1000, 2000 and 3000 tournaments (the 1000 and 2000 playout tournaments with a statistically significant difference with respect to the $\epsilon$-greedy strategies), and {\bf MCTS with NS$(0.4, 0.33, 0.0)$} won the 4000 and 5000 (but with a difference that is not statistically significant). 

Looking at the left-hand side of Figure \ref{fig:round-robin} closely, we can see that except for the extreme case of budget 500, na\"{i}ve sampling dominates the tournaments for low computation budgets (1000, 2000), and as the computation budget increases, the other strategies catch up. The exception is the case of budget 500, where $\epsilon$-greedy seems to work very well. After close inspection of the results, we noticed that this could be caused by two separate facts. First, the exploration constant used by  $\epsilon$-greedy ($\epsilon = 0.25$) was better suited for this low-computation budget setting than the higher exploration constants used by na\"{i}ve sampling in our experiments. We set all of these values experimentally based on overall performance across all tournaments. Second, we noticed that the performance gain of na\"{i}ve sampling with respect to $\epsilon$-greedy was smaller when using MCTS than when using MC, and in the special case of budget 500, MCTS with na\"{i}ve sampling seems to work very poorly. This makes us formulate the hypothesis that for very low computation budgets, the estimation of the rewards of the individual unit-actions made by the local MABs is not reliable, and thus, it does not help the search process. We will verify this hypothesis in the next section. 

Another interesting result we observe from Figure \ref{fig:round-robin} is that MC dominates MCTS for low computation budgets (all the MC bots are displayed with dashed lines, and the MCTS bots with solid lines), but as the computation budget increases, MCTS outperforms MC. This is observed both for $\epsilon$-greedy and for na\"{i}ve sampling. We would like to point out, however, that we set the exploration constants of the sampling strategies based on the performance of the MC AIs. Thus, it is possible that different exploration constants could make the MCTS AIs perform better. 

Finally, UCB1-FPU performed very poorly in this experiment, as in our previous experiments. Specially in the 12x12 maps. In the 8x8 maps, the performance was closer to the other policies, since the branching factors were smaller. Although FPU helped UCB1 significantly in our preliminary experiments (without adding an FPU constant, UCB1's performance is even lower), we found that finding the correct value for the FPU constant was not trivial. In the results presented in Figure \ref{fig:cmab1}, we fine-tuned this constant for each individual CMAB. But in real play, the FPU constant needs to be set automatically for each game situation. A fixed FPU constant seemed not to work, and we employed the dynamic scheme $\mathit{FPU} = f \times \rho(s) + (1-f)$ described above, which worked the best in our experiments, but still underperforms compared to the other strategies. However, comparing the results from Figures \ref{fig:cmab1} and \ref{fig:round-robin}, we believe there is still room for improvement with better FPU constant setting procedures.

In order to get better insight into these results, the center and right plots of Figure~\ref{fig:round-robin} show these results considering only the 8x8 maps (center) and only the 12x12 maps (right). Consistent with the results reported earlier in this paper, for the maps with larger branching factors, na\"{i}ve sampling performs better. Moreover, we can see that the crossover point between MC and MCTS occurs later for maps with larger branching factors.


\begin{figure}[t!]
    \centering
    \includegraphics[width=0.45\columnwidth]{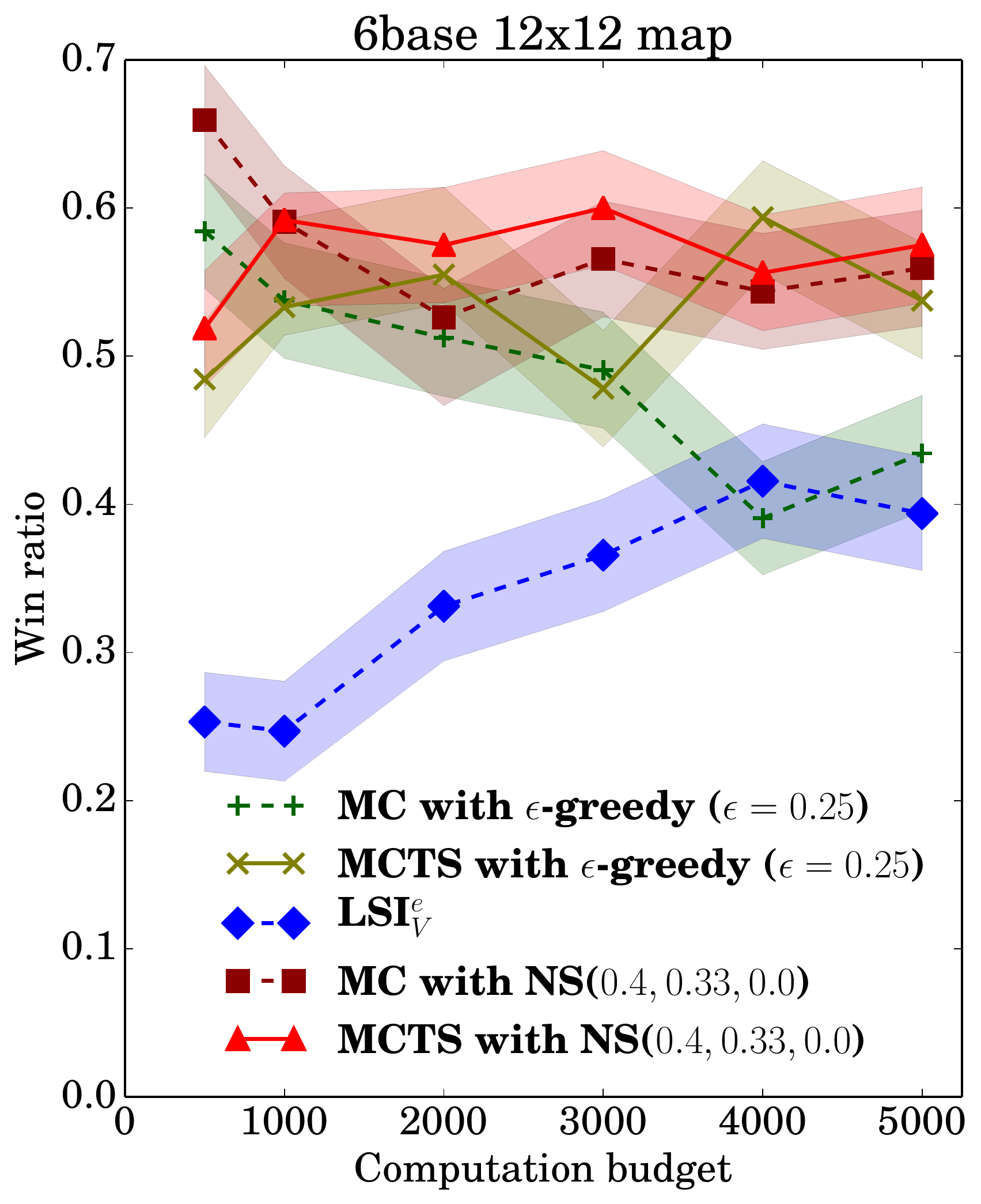}
    \caption{Win ratio of different approaches when doing a round-robin tournament with different computation budgets on a complex 12x12 map, where each player starts with six bases and six workers.}
    \label{fig:round-robin-large}
\end{figure}

Finally, in order to confirm the advantage of na\"{i}ve sampling over the other sampling strategies on larger maps, we performed an additional set of experiments on a more complex map (12x12 map, where each player started with 6 bases and 6 workers). Results are shown in Figure \ref{fig:round-robin-large}. Since this figure shows results for only one map, in order to reduce uncertainty, we made each AI play 40 times against each other, instead of 20 as in the previous plots. The results show that both na\"{i}ve sampling AIs consistently achieve higher win ratios than the other AIs, with the only other AI with comparable (but lower in most cases) performance being MCTS with $\epsilon$-greedy. In these experiments, the performance of MC with $\epsilon$-greedy degraded quickly compared with the other strategies. Moreover, given that the results shown in Figure \ref{fig:round-robin-large} correspond to only one map, we see a larger overlap of the 95\% confidence intervals in the figure, meaning that not all differences are statistically significant. Moreover UCB1-FPU is not shown since it achieved a win ratio of 0.

In conclusion, we can see that different strategies are better suited for different settings: when the computational budget is very low, MC seems to outperform MCTS, while as the computation budget increases this is reversed. Also, for low computational budgets $\epsilon$-greedy seems to achieve very good results (the next experiment will look at this more closely), and as the computation budget increases, na\"{i}ve sampling dominates (especially for situations with large branching factors). Finally, as the computation budget increases even further, then the result from different sampling strategies seems to converge ($\epsilon$-greedy, LSI and na\"{i}ve sampling using MC tend to converge to the same value, and $\epsilon$-greedy and na\"{i}ve sampling using MCTS also tend to converge).

\subsection{Experiment 5: Round-Robin with Small Budget}\label{sec:experiment5}

\begin{figure}[t!]
    \centering
    \includegraphics[width=0.48\columnwidth]{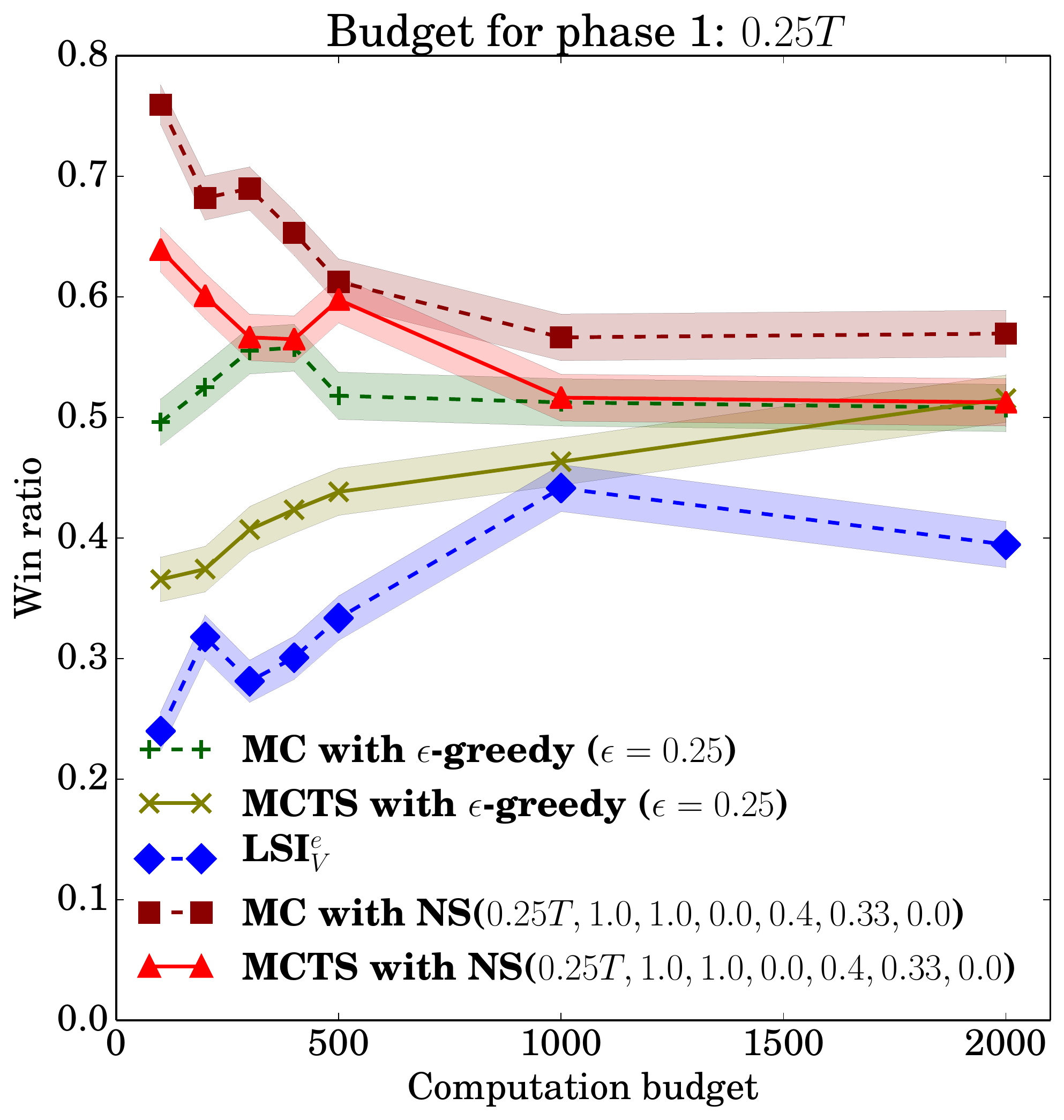}
    \includegraphics[width=0.48\columnwidth]{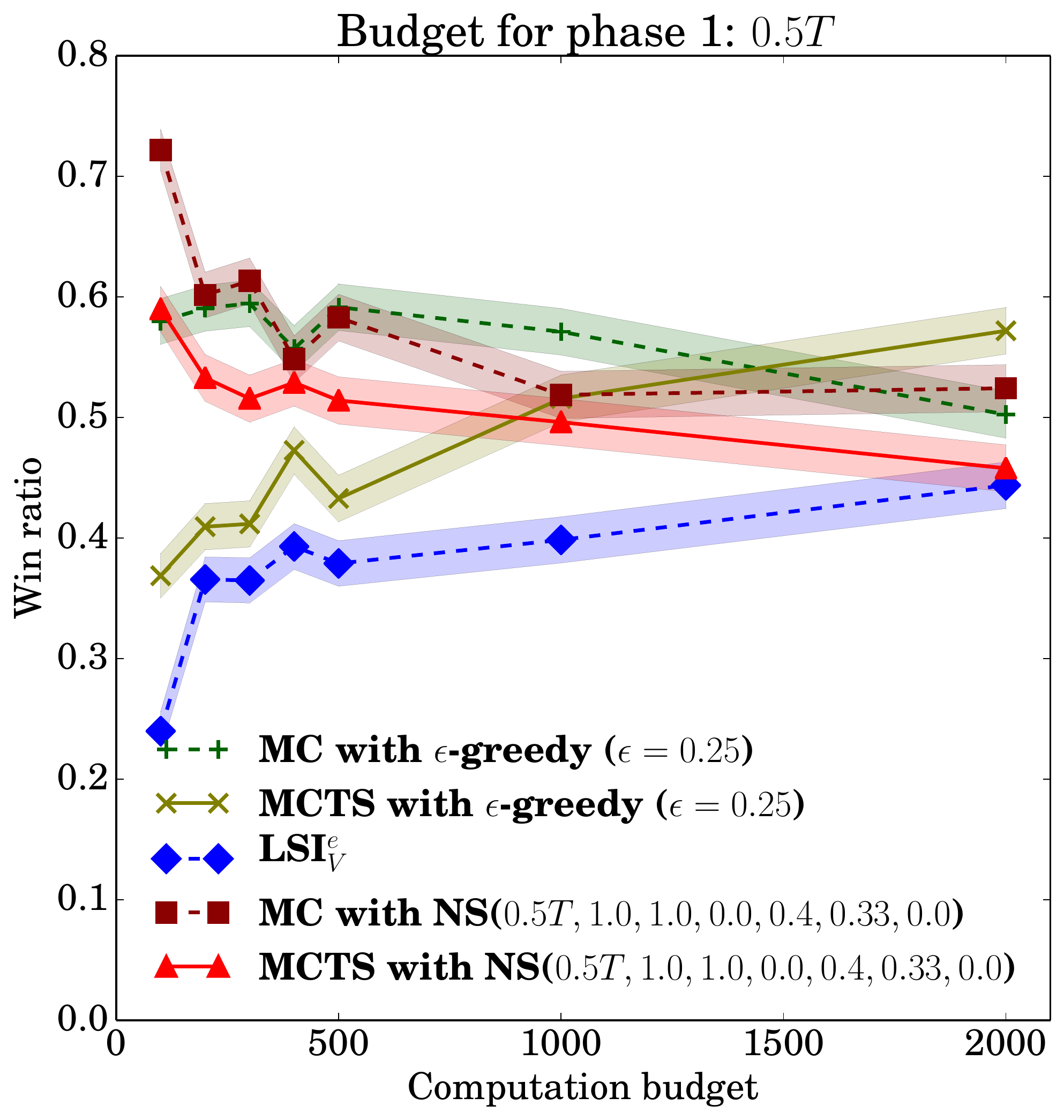}
    \caption{Win ratio of different approaches when doing a round-robin tournament, using two-phase na\"{i}ve sampling with low computation budgets averaged over all maps.}
    \label{fig:round-robin-2p}
\end{figure}

The previous experiment showed that an $\epsilon$-greedy strategy combined with MC seemed to work very well for very low computation budgets (500 playouts per decision cycle). Upon close inspection of the results, we observed that na\"{i}ve sampling struggled with low computation budgets, since in the first iterations, the local MABs still do not have meaningful estimations, and thus cannot guide exploration. In order to verify this hypothesis, we experimented with two-phase na\"{i}ve sampling strategies that would spend initially more time exploring (in order to get a good estimation in the local MABs) and only then start exploiting. 
Specifically, we repeated the round robin experiments, focusing on computation budgets between 100 to 2000, removing the {\bf UCB1-FPU} bot which did not perform well in the previous experiment, and using the following versions of na\"{i}ve sampling:
\begin{itemize}
\item {\bf NS}$(0.25T, 1.0, 1.0, 0.0, 0.4, 0.33, 0.0)$: during the first 25\% of the computation budget, this strategy will basically select macro-arms randomly, and then during the remaining 75\% of the computation budget, it will run the same parameter configuration of na\"{i}ve sampling as used in Experiment 4. 
\item {\bf NS}$(0.5T, 1.0, 1.0, 0.0, 0.4, 0.33, 0.0)$: the same, but exploring randomly during the first 50\% of the computation budget.
\end{itemize}

Figure \ref{fig:round-robin-2p} shows the experimental results. The first thing we see is that using 25\% of the computation budget to explore random macro-arms (left-hand side of Figure~\ref{fig:round-robin-2p}) seems to work much better than using 50\% of the computation budget (right-hand side of Figure~\ref{fig:round-robin-2p}). For example, using MC with a two-phase NS strategy achieves a win ratio of over 60\% with computation budgets lower than 500 in the 25\% exploration setting. This shows that when using very low computation budgets with na\"{i}ve sampling, it is important to dedicate some amount of the computation budget initially for exploration.

We performed preliminary experiments with larger computation budgets (not reported) and observed that for higher computation budgets, two-phase strategies either did not seem to make a difference or performed worse than standard na\"{i}ve sampling. So, two-phase sampling seems to only help in cases of low computation budgets.

\section{Related Work}\label{sec:related}

Several areas are related to the work presented in this paper: combinatorial multi-armed bandits (CMABs), AI techniques for RTS games, as well as more general work on multiagent planning or decentralized sequential decision making. While existing work on CMABs is covered in Section \ref{sec:other-cmab-strategies}, this section briefly discusses work on the other related areas, as well as their connection with the work presented in this paper.

Since the first call for research on RTS game AI by \citeauthor{buro2003rts}~\citeyear{buro2003rts}, a wide range of AI techniques have been explored to play RTS games. For example, reinforcement learning (RL) has been used for controlling individual units \cite{marthi2005writing,jaidee2012classq}, groups of units \cite{wender2012applying,2016arXiv160902993U}, and even to make high-level decisions in RTS games \cite{sharma2007transfer}. The main issue when deploying RL in RTS games is computational complexity, as the state and action space are very large. The aforementioned techniques address these problems by either focusing on individual units, small-scale combat, or by using domain knowledge to abstract the game state in order to simplify the problem. Although recent approaches are starting to scale up to larger and larger combat situations \cite{2016arXiv160902993U} by using techniques such as deep reinforcement learning, they are still far from scaling all the way up to full-game play. 

Other machine learning techniques, such as case-based approaches for learning to select among high-level strategies \cite{aha2005learning}, or for learning to choose the right ``build-order'' \cite{weber2009case}, have also been proposed. In our previous work, we showed that {\em learning from demonstration} \cite{ontanon2007case,Ontanon2010online}, although promising, also struggles to generalize due to the large variety of situations that can arise in RTS games.

More related to the work presented in this paper, there has also been a significant amount of work to design game-tree search approaches that can handle RTS games. Early work used plain Monte Carlo search \cite{ChungBS05}, but very soon, work shifted to Monte Carlo Tree Search \cite{balla2009uct}. And, as mentioned in Section \ref{subsec:mcts}, techniques now exist that can perform game tree search in domains with durative actions \cite{churchill2012abcd}, or simultaneous moves \cite{KovarskyB05heuristic,SaffidineFinnssonBuro2012AAAI}, both features of RTS games. Moreover, most recent work has focused on addressing the problem of the large branching factors present in RTS games. While in this paper we focused on a bandit strategy that can handle the combinatorial branching factor on RTS games, other work to address this problem can be categorized around four main lines:
\begin{itemize}
\item {\em Game state abstractions}: the idea is to re-represent the game state, removing some of the low-level details, in order to make the search space smaller. For example, Balla and Fern \citeyear{balla2009uct} represented the game state by clustering units into groups, idea which has been expanded upon in later work \cite{justesen2014script,uriarte2014game}.
\item {\em Portfolio approaches}: rather than re-representing the game state, portfolio approaches reduce the combinatorial branching factor of RTS games by only letting the AI pick amongst a predefined fixed set of scripts, rather than having to select among all the possible low-level actions. The first such work was proposed by \citeauthor{ChungBS05}~\citeyear{ChungBS05}, but modern versions of the approach perform either greedy search \cite{churchill2013portfolio} or MCTS \cite{justesen2014script}.
\item {\em Hierarchical search}: another idea that has been explored recently is that of performing search at several levels of abstraction, considering high-level decisions first, which condition the potential number of low-level decisions that can be taken. An example of this approach is the idea of {\em Adversarial HTN planning} \cite{ontanon2015adversarial}, which combines minimax search with HTN planning. Another example is the work of \citeauthor{stanescu2014hierarchical}~\citeyear{stanescu2014hierarchical}, who perform game tree search at two separate levels of abstraction, one informing the other.
\item Finally, a recent line of work, inspired by the success of {\em AlphaGO} \cite{silver2016mastering} has started to explore the idea of integrating machine learning into game tree search. For example, \citeauthor{stanescu2016evaluating}~\citeyear{stanescu2016evaluating} trained deep neural networks to automatically learn evaluation functions in $\mu$RTS, showing they could significantly outperform the base evaluation function. Another example is our previous work \cite{ontanon2016alpharts}, where we employed a very similar approach to {\em AlphaGO}, but using Bayesian models rather than neural networks, in order to inform the search in MCTS. 
\end{itemize}

Notice, however, that all four of these lines of work are orthogonal to the work presented in this paper, and, as a matter of fact, some of these strategies have been explored in conjunction with na\"{i}ve sampling.

Another related area is that of {\em multiagent planning} \cite{durfee2001distributed,de2005multi,brafman2008one}, which focuses on the problem of automated planning in domains where there are more than one agent performing actions. Of particular relevance to the work presented in this paper is the setting where planning is centralized and only execution is decentralized. A common framework to model this setting is that of {\em decentralized Markov Decision Processes} (DEC-MDPs and DEC-POMDPs) \cite{oliehoek2016concise}. A DEC-POMDP is a Partially Observable Markov Decision Process (POMDP), where there is more than one agent that can act simultaneously, each of them with their own partial perception of the world. All the agents in a DEC-POMDP attempt to maximize the same reward function. The particular case where the joint perception of all the agents corresponds to the complete state (i.e., when considering the joint perception makes the problem fully-observable) is called a DEC-MDP. Algorithms to address DEC-MDPs and DEC-POMDPs have been proposed in the literature, from systematic search, to dynamic programming \cite{hansen2004dynamic}, or approximate algorithms such as best-response approaches~\cite{nair2003taming} (which fix the policy of all agents but one, and iteratively each agent optimizes her policy with respect of the fixed policy of the others). The specific RTS-games setting we consider in this paper corresponds to a generalization of DEC-POMDPs, where different agents have different reward functions called {\em partially observable stochastic games} (POSG) \cite{hansen2004dynamic} (specifically, in our setting, all the units in an RTS game are divided among the two players, each player having her own reward function). The na\"{i}ve assumption made by na\"{i}ve sampling is also connected to the related idea of factored MDPs, where agents approximate the global reward function as a linear combination of local value functions \cite{guestrin2001multiagent}.

\section{Conclusions}\label{sec:conclusions}

Real-time strategy (RTS) games pose a significant challenge to game tree search approaches due to the very large branching factors they involve. In this paper, we explored the possibility of modeling RTS game situations as {\em combinatorial multi-armed bandit} (CMAB) problems, and study the theoretical and practical behavior of a new family of sampling strategies called na\"{i}ve sampling. We compare these sampling strategies against other sampling strategies in the literature for CMABs in the context of $\mu$RTS.

As our results indicate, for situations with small branching factors, na\"{i}ve sampling performs similar to other sampling strategies such as LSI or $\epsilon$-greedy. However, as the branching factor grows, the performance of other strategies degrades compared to na\"{i}ve sampling, especially under tight computation budgets, which is especially relevant for real-time games (as showed in Section 8.2, computation budgets in the order 1000 playouts per game cycle are to be expected).

As part of our future work we would like to better understand the behavior of na\"{i}ve sampling in the context of MCTS, where the computation budget that can be used in the inner nodes of the tree is smaller the deeper the node is in the tree. As the results in Experiments 4 and 5 showed, for very low computation budgets, the estimations performed by the local MABs might not be accurate enough to be reliable, and thus, two-phase strategies are more appropriate. However, how to decide when to use a two-phase versus a single-phase strategy is still unclear. Additionally, we are currently looking at sampling strategies that incorporate prior knowledge of the domain (such as the neural network models used by AlphaGO, see \citeR{silver2016mastering}) into na\"{i}ve sampling strategies (initial results on this direction indicate that significant gains can be achieved, see \citeR{ontanon2016alpharts}). We would also like to investigate better sampling strategies for CMABs by studying the relation of CMABs with other combinatorial optimization problems, and by studying the relation of na\"{i}ve sampling to sampling policies for continuous-valued bandits. Finally, we would like to apply na\"{i}ve sampling-based MCTS approaches to address large-scale RTS games such as {\em StarCraft}.

\appendix
\section{}\label{sec:appendix}

This section contains proofs to the propositions presented earlier, preceded by the simple regret analysis of $\epsilon$-greedy, which will set the basis for some of the proofs.

\subsection{Simple Regret Analysis of $\epsilon$-greedy}

Assume a MAB with $K$ arms, i.e., a single variable $X$ that can take values $\{v_1, ..., v_K\}$, and where $\overline{\mu}_t(v_i)$ is the average reward estimated for arm $i$ at iteration $t$.

Assume also that the expected difference in expected reward between the optimal arm and a non-optimal arm is $D =  E_{v_i \in \mathcal{X}^-} (\mu^* - \mu(v_i))$, where $\mu^*$ is the expected reward of an optimal arm, and $\mathcal{X}^- = \{v_i \in \mathcal{X} | \mu(v_i) < \mu^*\}$ is the set of non-optimal arms; and that the minimum difference between the optimal arm and a non-optimal arm is at least $d = min_{v_i \in \mathcal{X}^-} (\mu^* - \mu(v_i))$.

Given a suboptimal arm $v_i$ and an optimal arm $v^*$ that have been sampled at least $m$ times, the probability that $\overline{\mu}_t(v_i) \geq \overline{\mu}_t(v^*)$ can be bounded in the following way.
By Hoeffding's inequality~\cite{hoeffding1963probability}, we know that the probability that the empirical estimation $\overline{S}$ of the mean of a variable $S$ differs in more than $d$ from the actual mean $E[\overline{S}]$ after having sampled it $m$ times, is bounded by $P(E[\overline{X}] - \overline{S} \geq d) \leq e^{-2d^2m}$. Now let $S$ be the difference in observed reward between $v^*$ and $v_i$. Then the empirical estimate $\overline{S}$ is $(\overline{\mu}_t(v^*) - \overline{\mu}_t(v_i))$, and by assumption we know that $E[\overline{S}] = \Delta \geq d$. Thus, by Hoeffding's inequality, we know that:
$$P(\Delta - (\overline{\mu}_t(v^*) - \overline{\mu}_t(v_i)) \geq \Delta) \,\, \leq \,\, e^{-2\Delta^2m} \,\, \leq \,\, e^{-2d^2m}$$


And thus:
\[
\begin{array}{rcl}
P(\overline{\mu}_t(v_i) \geq \overline{\mu}_t(v^*)) & = &  P((\overline{\mu}_t(v^*) - \overline{\mu}_t(v_i) \leq 0) \\
& = & P(d - (\overline{\mu}_t(v^*) - \overline{\mu}_t(v_i)) \geq d) \\
& \leq & e^{-2d^2m}
\end{array}
\]

Let $v_t^{best}$ be the arm believed to be the best at iteration $t$ (the one with the highest estimated reward so far), $v_t^{best-no}$ be the arm from the set of non-optimal arms $\mathcal{X}^-$ that at iteration $t$ has the highest estimated reward so far, and $n_t^{best}$ and $n_t^{best-no}$ the number of times these arms have been sampled respectively. Now, let us assume that there is a single optimal arm $v^*$ (which has been sampled $n_t^*$ times). Now, notice that if ${\mu}_t(v_t^{best-no}) > \overline{\mu}_t(v^*)$, then 
$v_t^{best}$ will be $v_t^{best-no}$ instead of $v^*$, and thus the arm believed to be the best at iteration $t$ is not the optimal one. Therefore, we can estimate the probability of $v_t^{best}$ not to be the optimal arm as:

\[
\begin{array}{rcl}
P(\overline{\mu}_t(v_t^{best-no}) \geq \overline{\mu}_t(v^*)) & = & P((\overline{\mu}_t(v^*) - \overline{\mu}_t(v_t^{best-no})) \leq 0) \\
& \leq & e^{-2d^2 min(n_t^{best-no},n_t^*)} \\
& \leq & e^{-2d^2 \frac{\epsilon}{K} t} \\
\end{array}
\]


If we lift the assumption that there is a single optimal arm, then the probability of $v_t^{best}$ not to be optimal is even lower, and thus this bound still applies. Thus, the probability of choosing the wrong arm after $t$ iterations decreases exponentially. The regret for $\epsilon$-greedy is:
\begin{itemize}
\item {\em Instantaneous regret}:
	$$r'_t \approx \left(\epsilon\frac{K^{no}}{K} + (1-\epsilon)e^{-2d^2 \frac{\epsilon}{K} t}\right)D$$
\noindent where $K^{no} < K$ is the number of non-optimal arms. Thus, when $t \to \infty$, this tends to $\epsilon\frac{K^{no}}{K}D$, which is a constant; and when $K$ is very large (and $K^{no}/K \approx 1$), can be simplified as $\epsilon D$.
	
\item {\em Cumulative regret}: since the instantaneous regret tends to a constant, cumulative regret is linear $R_t = O(t\epsilon D)$
\item {\em Simple regret}: the simple regret decreases exponentially:
	$$ r_t \approx D e^{-2d^2 \frac{\epsilon}{K} t} $$
\end{itemize}

Moreover, notice that decreasing $\epsilon$ reduces the instantaneous and cumulative regret but increases the simple regret. $\epsilon = 1$, which achieves the highest expected instantaneous and cumulative regret, achieves the lowest expected simple regret.

\subsection{Propositions and Proofs}

{\em Proposition \ref{ens-cumulative-proposition}:}
The {\em cumulative regret} of NS($\epsilon_0$,$\epsilon_l$,$\epsilon_g$) grows linearly as $R_T = O((1 - p^*) D T)$, where $T$ is the number of iterations, $D$ is the expected difference in expected reward between an optimal macro-arm and a non-optimal macro-arm, and $p^* \geq (1 - \epsilon_0)(1 - \epsilon_g)$ is the probability of selecting an optimal macro-arm when $T \to \infty$.

{\em Proof:}
Formally, $D =  E_{V_i \in \mathcal{X}^-} (\mu^* - \mu(V_i))$, where $\mathcal{X}^- = \{V_i \in \mathcal{X} | L(V_i) \wedge \mu(V_i) < \mu^*\}$ is the set of legal non-optimal macro-arms.

The probability of selecting an optimal macro-arm $V^*$ when $T \to \infty$ can be calculated as follows. When $T \to \infty$, we can assume that each macro-arm has been sampled enough times as for $V^*$ having higher reward than all the non-optimal macro-arms. Thus, NS($\epsilon_0$,$\epsilon_l$,$\epsilon_g$) would select it in the following circumstances:
\begin{itemize}
\item When exploiting, $V^*$ will be selected with probability $(1-\epsilon_g) + \epsilon_g/N$ if there is a single optimal macro-arm, and higher if there is more than one optimal macro-arm (where $N$ is the number of legal macro-arms). Assuming that $N$ is large, this is approximately $(1 - \epsilon_g)$.
\item When exploring, $V^*$ will be selected with probability at least $\prod_{i = 1...n} ((1-\epsilon_l) + \epsilon_l/K_i)$ (the probability will be higher if there is more than one optimal macro arm). 
\end{itemize}
Thus, for large $N$ the probability of selecting an optimal macro-arm $V^*$ is at least: $p^* = \epsilon_0\left(\prod_{i = 1...n} ((1-\epsilon_l) + \epsilon_l/K_i)\right) + (1-\epsilon_0)(1-\epsilon_g)$, from which we know that $p^* \geq (1-\epsilon_0)(1-\epsilon_g)$ (and if the number of variables $n$ is large, $\epsilon_l > 0$, and there are few optimal macro-arms, the inequality will be tight).

Thus, the instantaneous regret of NS($\epsilon_0$,$\epsilon_l$,$\epsilon_g$) will be $r_T' \approx (1-p^*)D$, which implies that the cumulative regret will be $r_T = O((1-p^*)DT)$.
\hfill $\square$

{\em Proposition \ref{ens-simple-proposition}:}
The {\em simple regret} of NS($\epsilon_0$,$\epsilon_l$,$\epsilon_g$) decreases at an exponential rate as $r_T = O(D e^{-2d^2 T p_i})$, where $p_i \geq \epsilon_0 \Pi_{j = 1...n}\frac{\epsilon_l}{K_j} + (1-\epsilon_0)\frac{\epsilon_g}{N}$, $D$ is as in Proposition~\ref{ens-cumulative-proposition}, and $d$ is the minimum difference between an optimal macro-arm and a non-optimal macro-arm.

\begin{proof}
Formally, $d = min_{V_i \in \mathcal{X}^-} (\mu^* - \mu(V_i))$, where $\mathcal{X}^-$ is as in Proposition \ref{ens-cumulative-proposition}. Given a suboptimal macro-arm $V_i$ and an optimal macro-arm $V^*$, after a sufficiently large number of iterations $T$: 
\begin{itemize}
\item The probability of selecting $V_i$ is $p_i \geq \epsilon_0 \Pi_{j = 1...n}\frac{\epsilon_l}{K_j} + (1-\epsilon_0)\frac{\epsilon_g}{N}$, and thus, we can expect the arm to have been selected at least $n_T = T p_i$ times.
\item The probability of selecting an optimal macro-arm $V^*$ is $p^* \geq (1 - \epsilon_0)(1 - \epsilon_g)$, and thus, we can expect the arm to have been selected at least $n_T^* = T p^*$ times.
\end{itemize}
Now, given $V_t^{best}$ to be the best macro-arm at iteration $t$, the probability that $V_t^{best}$ is not an optimal arm is (using Hoeffding's inequality, see~\citeR{hoeffding1963probability}):
\[
\begin{array}{rcl}
P(\overline{\mu}_t(V_t^{best}) \geq \overline{\mu}_t(V^*)) & = & P((\overline{\mu}_t(V^*) - \overline{\mu}_t(V_t^{best})) \leq 0) \\
& \leq & e^{-2d^2 min(n_t,n_t^*)} 
\end{array}
\]

Assuming $\epsilon_0 \leq 0.5$, $\epsilon_g \leq 0.5$ and $\epsilon_l \leq 0.5$, and $N \gg 2$, $p_i$ is expected to be lower than $p^*$, and thus:
$$P((\overline{\mu}_t(V^*) - \overline{\mu}_t(V_t^{best})) \leq 0) \leq e^{-2d^2 n_t} $$
Thus, if the expected difference in expected reward between the optimal arm and a non-optimal arm is $D$, after a sufficiently large number of iterations $T$, the expected simple regret $r_T \approx D e^{-2d^2 n_T}$.
\end{proof}

{\em Proposition \ref{prop:2phase-explore}:}
In a CMAB with $n$ variables, the probability that after $t$ iterations using a NS$(\epsilon_0, \epsilon_l, \epsilon_g)$ sampling strategy an optimal macro-arm $V^*$ has not been explored at least once, decreases exponentially as a function of $t$, and is at most $(1-p)^{t\epsilon_0}$, where $p = \prod_{i = 1 ... n} \left(\epsilon_l/K_i\right)$.

\begin{proof}
The expected number of exploration iterations done after $t$ iterations is $m = t\epsilon_0$. When exploring, the probability that a given value $v_i^j$ is selected for variable $X_i$ at a given exploration iteration, is at least $\epsilon_l/K_i$ (it could be higher, if $v_i^j$ happens to be the value with the highest expected reward so far). Thus, the probability of selecting $V^* = (v_1^*, ..., v_n^*)$ during exploration is at least $p = \prod_{i = 1 ... n} \left(\epsilon_l/K_i\right)$. Therefore, the probability of not selecting $V^*$ after $m$ exploration iterations is at most $(1-p)^m$, i.e., it decreases exponentially as a function of $t$. 
\end{proof}

{\em Proposition \ref{prop:2phase-cumulative}:}
The {\em cumulative regret} of NS($k,\epsilon_0^1,\epsilon_l^1,\epsilon_g^1, \epsilon_0^2, 0, \epsilon_g^2$) when $k$ is a constant grows linearly when $T \gg k$: 
$$R_T = O\left(t\left[ (1 -\epsilon_g - q_k + \epsilon_g q_k)d + \epsilon_g D \right]\right)$$
\noindent where $q_k = 1 - (1-p)^{k\epsilon_0^1}$, and $p = \prod_{i = 1 ... n} \left(\epsilon_l^1/K_i\right)$.

\begin{proof}
Let us use $D$ and $d$ as before (difference between an optimal macro-arm and a random macro-arm, and difference between the best non-optimal macro-arm and an optimal macro-arm respectively). Notice that if $\epsilon_0^2 = 0$, it means that the second stage is just an $\epsilon$-greedy strategy with the set of explored macro-arms with parameter $\epsilon_g$. Additionally, according to Proposition \ref{prop:2phase-explore}, with probability at least $q_k = 1 - (1-p)^{k\epsilon_0^1}$, an optimal macro-arm will be explored during the first phase. 

Therefore, with probability $q_k$, for $T \gg k$, an optimal macro-arm will be in the global MAB. Also, if $T \gg k$, the number of times the macro-arms in the global MAB have been sampled during the first $k$ iterations will be negligible. Thus we can just focus on the second stage. Therefore, with probability $q_k$ we will have a cumulative regret $O(t\epsilon_g D)$.

On the other hand, with probability $(1-q_k)$, the instantaneous regret will be at least $d$ (since no optimal macro-arm is in the global MAB, and thus, the difference between the best macro-arm in the global MAB and an optimal macro-arm is at least $d$). In this case, instantaneous regret will converge to a higher constant: $r'_t \approx aD + (1-a)d$, where $a = \left(\epsilon_g\frac{K^{no}}{K} + (1-\epsilon_g)e^{-2d^2 \frac{\epsilon_g}{K} t}\right)$. When $t \to \infty$, and $K$ is large, this converges to: $r'_t \approx d + \epsilon_g (D-d)$, which gives a cumulative regret $O\left(t(d + \epsilon_g (D-d))\right)$

Thus, the expected cumulative regret of two-phase na\"{i}ve when $k$ is a constant and $\epsilon_l^0 = 0$ is (after rearranging):
$$R_t = O\left(t\left[ (1 -\epsilon_g - q_k + \epsilon_g q_k)d + \epsilon_g D \right]\right)$$
\end{proof}

{\em Proposition \ref{prop:2phase-simple}:}
The {\em simple regret} of NS($k,\epsilon_0^1,\epsilon_l^1,\epsilon_g^1, \epsilon_0^2, 0, \epsilon_g^2$) when $k$ is a constant and $T \gg k$ is lower bounded by $(1-q_k)d$, where $q_k = 1 - (1-p)^{k\epsilon_0^1}$, and $d$ is difference between the best non-optimal macro-arm and an optimal macro-arm.

\begin{proof}
Following a similar line of reasoning as for Proposition \ref{prop:2phase-cumulative}, with probability $q_k$ at least one optimal macro-arm will be in the global MAB and for $T \gg k$ we can just focus on the second stage. Thus, with probability $q_k$, following the $\epsilon$-greedy analysis presented above, the simple regret will be approximately $D e^{-2d^2 \frac{\epsilon_g}{K} T}$.

With probability $(1-q_k)$, however, there will be no optimal macro-arm in the global MAB. Let $d'$ be the difference in reward between the best macro-arm in the global MAB ($V^+$) and the second best, and $D'$ the average difference in reward between the best macro-arm in the global MAB and the rest. In this case, the simple regret with respect of not choosing $V^+$ will be approximately $D' e^{-2d'^2 \frac{\epsilon_g}{K} T}$. However, since the difference in reward between $V^+$ and an optimal macro arm is at least $d$, then we need to add $d$ to the actual simple regret.

Thus, we have that the simple regret of two-phase na\"{i}ve when $k$ is a constant and $\epsilon_l^0 = 0$ is:
$$r_t \approx q_k \left(D e^{-2d^2 \frac{\epsilon_g}{K} T}\right) + (1-q_k)\left(d + D' e^{-2d'^2 \frac{\epsilon_g}{K} T}\right)$$
Which, when $T \to \infty$, converges to: $(1-q_k)d$.
\end{proof}

\bibliographystyle{theapa}                                                    
\bibliography{references}

\begin{thebibliography}{}

\bibitem[\protect\BCAY{Aha, Molineaux,\ \BBA\ Ponsen}{Aha
  et~al.}{2005}]{aha2005learning}
Aha, D.~W., Molineaux, M., \BBA\ Ponsen, M. \BBOP2005\BBCP.
\newblock \BBOQ Learning to win: Case-based plan selection in a real-time
  strategy game\BBCQ\
\newblock In {\Bem Case-based reasoning research and development}, \BPGS\
  5--20. Springer.

\bibitem[\protect\BCAY{Auer, Cesa-Bianchi,\ \BBA\ Fischer}{Auer
  et~al.}{2002}]{auer2002finite}
Auer, P., Cesa-Bianchi, N., \BBA\ Fischer, P. \BBOP2002\BBCP.
\newblock \BBOQ Finite-time analysis of the multiarmed bandit problem\BBCQ\
\newblock {\Bem Machine learning}, {\Bem 47\/}(2), 235--256.

\bibitem[\protect\BCAY{Balla\ \BBA\ Fern}{Balla\ \BBA\
  Fern}{2009}]{balla2009uct}
Balla, R.-K.\BBACOMMA\  \BBA\ Fern, A. \BBOP2009\BBCP.
\newblock \BBOQ {UCT} for tactical assault planning in real-time strategy
  games\BBCQ\
\newblock In {\Bem Proceedings of the International Joint Conference on
  Artificial Intelligence (IJCAI 2009)}, \BPGS\ 40--45.

\bibitem[\protect\BCAY{Brafman\ \BBA\ Domshlak}{Brafman\ \BBA\
  Domshlak}{2008}]{brafman2008one}
Brafman, R.~I.\BBACOMMA\  \BBA\ Domshlak, C. \BBOP2008\BBCP.
\newblock \BBOQ From one to many: Planning for loosely coupled multi-agent
  systems\BBCQ\
\newblock In {\Bem Proceedings of the International Conference on Automated
  Planning and Scheduling (ICAPS 2008)}, \BPGS\ 28--35.

\bibitem[\protect\BCAY{Browne, Powley, Whitehouse, Lucas, Cowling, Rohlfshagen,
  Tavener, Perez, Samothrakis,\ \BBA\ Colton}{Browne
  et~al.}{2012}]{browne2012survey}
Browne, C., Powley, E., Whitehouse, D., Lucas, S., Cowling, P., Rohlfshagen,
  P., Tavener, S., Perez, D., Samothrakis, S., \BBA\ Colton, S. \BBOP2012\BBCP.
\newblock \BBOQ A survey of {M}onte {C}arlo {T}ree {S}earch methods\BBCQ\
\newblock {\Bem IEEE Transactions on Computational Intelligence and {AI} in
  Games}, {\Bem 4\/}(1), 1--43.

\bibitem[\protect\BCAY{{Bubeck}\ \BBA\ {Cesa-Bianchi}}{{Bubeck}\ \BBA\
  {Cesa-Bianchi}}{2012}]{bubeck2012regret}
{Bubeck}, S.\BBACOMMA\  \BBA\ {Cesa-Bianchi}, N. \BBOP2012\BBCP.
\newblock \BBOQ Regret analysis of stochastic and nonstochastic multi-armed
  bandit problems\BBCQ\
\newblock {\Bem ArXiv e-prints}, {\Bem arXiv:1204.5721}.

\bibitem[\protect\BCAY{Bubeck, Munos,\ \BBA\ Stoltz}{Bubeck
  et~al.}{2011}]{bubeck2011pure}
Bubeck, S., Munos, R., \BBA\ Stoltz, G. \BBOP2011\BBCP.
\newblock \BBOQ Pure exploration in finitely-armed and continuous-armed
  bandits\BBCQ\
\newblock {\Bem Theoretical Computer Science}, {\Bem 412\/}(19), 1832--1852.

\bibitem[\protect\BCAY{Bubeck, Munos, Stoltz,\ \BBA\ Szepesvari}{Bubeck
  et~al.}{2008}]{bubeck2008online}
Bubeck, S., Munos, R., Stoltz, G., \BBA\ Szepesvari, C. \BBOP2008\BBCP.
\newblock \BBOQ Online optimization in $\mathcal{X}$-armed bandits\BBCQ\
\newblock In {\Bem Twenty-Second Annual Conference on Neural Information
  Processing Systems}.

\bibitem[\protect\BCAY{Buro}{Buro}{2003}]{buro2003rts}
Buro, M. \BBOP2003\BBCP.
\newblock \BBOQ Real-time strategy games: a new {AI} research challenge\BBCQ\
\newblock In {\Bem Proceedings of the International Joint Conference on
  Artificial Intelligence (IJCAI 2003)}, \BPGS\ 1534--1535, San Francisco, CA,
  USA. Morgan Kaufmann Publishers Inc.

\bibitem[\protect\BCAY{Cazenave}{Cazenave}{2015}]{cazenave2015sequential}
Cazenave, T. \BBOP2015\BBCP.
\newblock \BBOQ Sequential halving applied to trees\BBCQ\
\newblock {\Bem {IEEE} Transactions on Computational Intelligence and {AI} in
  Games}, {\Bem 7\/}(1), 102--105.

\bibitem[\protect\BCAY{Chen, Wang,\ \BBA\ Yuan}{Chen
  et~al.}{2013}]{chen2013combinatorial}
Chen, W., Wang, Y., \BBA\ Yuan, Y. \BBOP2013\BBCP.
\newblock \BBOQ Combinatorial multi-armed bandit: General framework and
  applications\BBCQ\
\newblock In {\Bem Proceedings of the 30th International Conference on Machine
  Learning (ICML 2013)}, \BPGS\ 151--159.

\bibitem[\protect\BCAY{Chinchalkar}{Chinchalkar}{1996}]{Chinchalkar1996upperbound}
Chinchalkar, S. \BBOP1996\BBCP.
\newblock \BBOQ An upper bound for the number of reachable positions\BBCQ\
\newblock {\Bem ICCA Journal}, {\Bem 19\/}(3).

\bibitem[\protect\BCAY{Chung, Buro,\ \BBA\ Schaeffer}{Chung
  et~al.}{2005}]{ChungBS05}
Chung, M., Buro, M., \BBA\ Schaeffer, J. \BBOP2005\BBCP.
\newblock \BBOQ Monte carlo planning in {RTS} games\BBCQ\
\newblock In {\Bem Proceedings of the IEEE Conference on Computational
  Intelligence in Games (CIG 2005)}.

\bibitem[\protect\BCAY{Churchill, Saffidine,\ \BBA\ Buro}{Churchill
  et~al.}{2012}]{churchill2012abcd}
Churchill, D., Saffidine, A., \BBA\ Buro, M. \BBOP2012\BBCP.
\newblock \BBOQ Fast heuristic search for {RTS} game combat scenarios\BBCQ\
\newblock In {\Bem Proceedings of the Artificial Intelligence and Interactive
  Digital Entertainment conference (AIIDE 2012)}. The AAAI Press.

\bibitem[\protect\BCAY{Churchill\ \BBA\ Buro}{Churchill\ \BBA\
  Buro}{2013}]{churchill2013portfolio}
Churchill, D.\BBACOMMA\  \BBA\ Buro, M. \BBOP2013\BBCP.
\newblock \BBOQ Portfolio greedy search and simulation for large-scale combat
  in {S}tar{C}raft\BBCQ\
\newblock In {\Bem IEEE Conference on Computational Intelligence in Games (CIG
  2013)}, \BPGS\ 1--8.

\bibitem[\protect\BCAY{De~Weerdt, Ter~Mors,\ \BBA\ Witteveen}{De~Weerdt
  et~al.}{2005}]{de2005multi}
De~Weerdt, M., Ter~Mors, A., \BBA\ Witteveen, C. \BBOP2005\BBCP.
\newblock \BBOQ Multi-agent planning: An introduction to planning and
  coordination\BBCQ\
\newblock In {\Bem Handouts of the European Agent Summer}.

\bibitem[\protect\BCAY{Durfee}{Durfee}{2001}]{durfee2001distributed}
Durfee, E.~H. \BBOP2001\BBCP.
\newblock \BBOQ Distributed problem solving and planning\BBCQ\
\newblock In {\Bem Multi-agent systems and applications}, \BPGS\ 118--149.
  Springer.

\bibitem[\protect\BCAY{Feldman\ \BBA\ Domshlak}{Feldman\ \BBA\
  Domshlak}{2014}]{feldman2014simple}
Feldman, Z.\BBACOMMA\  \BBA\ Domshlak, C. \BBOP2014\BBCP.
\newblock \BBOQ Simple regret optimization in online planning for {Markov}
  decision processes\BBCQ\
\newblock {\Bem Journal of Artificial Intelligence Research}, {\Bem 51},
  165--205.

\bibitem[\protect\BCAY{Gai, Krishnamachari,\ \BBA\ Jain}{Gai
  et~al.}{2010}]{gai2010learning}
Gai, Y., Krishnamachari, B., \BBA\ Jain, R. \BBOP2010\BBCP.
\newblock \BBOQ Learning multiuser channel allocations in cognitive radio
  networks: A combinatorial multi-armed bandit formulation\BBCQ\
\newblock In {\Bem 2010 IEEE Symposium on New Frontiers in Dynamic Spectrum},
  \BPGS\ 1--9.

\bibitem[\protect\BCAY{Gelly\ \BBA\ Silver}{Gelly\ \BBA\
  Silver}{2007}]{gelly2007combining}
Gelly, S.\BBACOMMA\  \BBA\ Silver, D. \BBOP2007\BBCP.
\newblock \BBOQ Combining online and offline knowledge in {UCT}\BBCQ\
\newblock In {\Bem Proceedings of the 24th International Conference on Machine
  learning (ICML 2007)}, \BPGS\ 273--280. ACM.

\bibitem[\protect\BCAY{Gelly\ \BBA\ Wang}{Gelly\ \BBA\
  Wang}{2006}]{gelly2006exploration}
Gelly, S.\BBACOMMA\  \BBA\ Wang, Y. \BBOP2006\BBCP.
\newblock \BBOQ Exploration exploitation in {Go}: {UCT} for monte-carlo
  {Go}\BBCQ\
\newblock In {\Bem NIPS: Neural Information Processing Systems Conference
  On-line trading of Exploration and Exploitation Workshop}.

\bibitem[\protect\BCAY{Guestrin, Koller,\ \BBA\ Parr}{Guestrin
  et~al.}{2001}]{guestrin2001multiagent}
Guestrin, C., Koller, D., \BBA\ Parr, R. \BBOP2001\BBCP.
\newblock \BBOQ Multiagent planning with factored {MDPs}\BBCQ\
\newblock In {\Bem Proceedings of the Conference on Neural Information
  Processing Systems (NIPS 2001)}, \lowercase{\BVOL}~1, \BPGS\ 1523--1530.

\bibitem[\protect\BCAY{Hansen, Bernstein,\ \BBA\ Zilberstein}{Hansen
  et~al.}{2004}]{hansen2004dynamic}
Hansen, E.~A., Bernstein, D.~S., \BBA\ Zilberstein, S. \BBOP2004\BBCP.
\newblock \BBOQ Dynamic programming for partially observable stochastic
  games\BBCQ\
\newblock In {\Bem Proceedings of the Association for the Advancement of
  Artificial Intelligence Conference (AAAI 204)}, \lowercase{\BVOL}~4, \BPGS\
  709--715.

\bibitem[\protect\BCAY{Hoeffding}{Hoeffding}{1963}]{hoeffding1963probability}
Hoeffding, W. \BBOP1963\BBCP.
\newblock \BBOQ Probability inequalities for sums of bounded random
  variables\BBCQ\
\newblock {\Bem Journal of the American statistical association}, {\Bem
  58\/}(301), 13--30.

\bibitem[\protect\BCAY{Jaidee\ \BBA\ Mu{\~n}oz-Avila}{Jaidee\ \BBA\
  Mu{\~n}oz-Avila}{2012}]{jaidee2012classq}
Jaidee, U.\BBACOMMA\  \BBA\ Mu{\~n}oz-Avila, H. \BBOP2012\BBCP.
\newblock \BBOQ Classq-l: A q-learning algorithm for adversarial real-time
  strategy games\BBCQ\
\newblock In {\Bem Eighth Artificial Intelligence and Interactive Digital
  Entertainment Conference}.

\bibitem[\protect\BCAY{Justesen, Tillman, Togelius,\ \BBA\ Risi}{Justesen
  et~al.}{2014}]{justesen2014script}
Justesen, N., Tillman, B., Togelius, J., \BBA\ Risi, S. \BBOP2014\BBCP.
\newblock \BBOQ Script-and cluster-based {UCT} for {StarCraft}\BBCQ\
\newblock In {\Bem Computational Intelligence and Games (CIG), 2014 IEEE
  Conference on}, \BPGS\ 1--8. IEEE.

\bibitem[\protect\BCAY{Karnin, Koren,\ \BBA\ Somekh}{Karnin
  et~al.}{2013}]{karnin2013almost}
Karnin, Z., Koren, T., \BBA\ Somekh, O. \BBOP2013\BBCP.
\newblock \BBOQ Almost optimal exploration in multi-armed bandits\BBCQ\
\newblock In {\Bem Proceedings of the 30th International Conference on Machine
  Learning (ICML-13)}, \BPGS\ 1238--1246.

\bibitem[\protect\BCAY{Kocsis\ \BBA\ Szepesv�ri}{Kocsis\ \BBA\
  Szepesv�ri}{2006}]{Kocsis06banditbased}
Kocsis, L.\BBACOMMA\  \BBA\ Szepesv�ri, C. \BBOP2006\BBCP.
\newblock \BBOQ Bandit based monte-carlo planning\BBCQ\
\newblock In {\Bem Proceedings of the European Conference on Machine Learning
  (ECML 2006)}, \BPGS\ 282--293. Springer.

\bibitem[\protect\BCAY{Kovarsky\ \BBA\ Buro}{Kovarsky\ \BBA\
  Buro}{2005}]{KovarskyB05heuristic}
Kovarsky, A.\BBACOMMA\  \BBA\ Buro, M. \BBOP2005\BBCP.
\newblock \BBOQ Heuristic search applied to abstract combat games\BBCQ\
\newblock In {\Bem Canadian Conference on AI}, \BPGS\ 66--78.

\bibitem[\protect\BCAY{Kuhn}{Kuhn}{1955}]{kuhn1955hungarian}
Kuhn, H.~W. \BBOP1955\BBCP.
\newblock \BBOQ The hungarian method for the assignment problem\BBCQ\
\newblock {\Bem Naval research logistics quarterly}, {\Bem 2\/}(1-2), 83--97.

\bibitem[\protect\BCAY{Marthi, Russell,\ \BBA\ Latham}{Marthi
  et~al.}{2005}]{marthi2005writing}
Marthi, B., Russell, S., \BBA\ Latham, D. \BBOP2005\BBCP.
\newblock \BBOQ Writing stratagus-playing agents in concurrent {ALisp}\BBCQ\
\newblock In {\Bem Proceedings of the International Joint Conference on
  Artificial Intelligence (IJCAI 2005)}.

\bibitem[\protect\BCAY{Nair, Tambe, Yokoo, Pynadath,\ \BBA\ Marsella}{Nair
  et~al.}{2003}]{nair2003taming}
Nair, R., Tambe, M., Yokoo, M., Pynadath, D., \BBA\ Marsella, S.
  \BBOP2003\BBCP.
\newblock \BBOQ Taming decentralized {POMDPs}: Towards efficient policy
  computation for multiagent settings\BBCQ\
\newblock In {\Bem Proceedings of the International Joint Conference on
  Artificial Intelligence (IJCAI 2003)}, \BPGS\ 705--711.

\bibitem[\protect\BCAY{Oliehoek\ \BBA\ Amato}{Oliehoek\ \BBA\
  Amato}{2016}]{oliehoek2016concise}
Oliehoek, F.~A.\BBACOMMA\  \BBA\ Amato, C. \BBOP2016\BBCP.
\newblock {\Bem Concise Introduction to Decentralized {POMDPs}}.
\newblock Springer.

\bibitem[\protect\BCAY{Onta{\~n}{\'o}n, Mishra, Sugandh,\ \BBA\
  Ram}{Onta{\~n}{\'o}n et~al.}{2010}]{Ontanon2010online}
Onta{\~n}{\'o}n, S., Mishra, K., Sugandh, N., \BBA\ Ram, A. \BBOP2010\BBCP.
\newblock \BBOQ On-line case-based planning\BBCQ\
\newblock {\Bem Computational Intelligence}, {\Bem 26\/}(1), 84--119.

\bibitem[\protect\BCAY{Onta{\~n}\'{o}n}{Onta{\~n}\'{o}n}{2013}]{ontanon2013combinatorial}
Onta{\~n}\'{o}n, S. \BBOP2013\BBCP.
\newblock \BBOQ The combinatorial multi-armed bandit problem and its
  application to real-time strategy games\BBCQ\
\newblock In {\Bem Proceedings of the AAAI Artificial Intelligence and
  Interactive Digital Entertainment conference (AIIDE 2013)}.

\bibitem[\protect\BCAY{Onta{\~n}{\'o}n}{Onta{\~n}{\'o}n}{2016}]{ontanon2016alpharts}
Onta{\~n}{\'o}n, S. \BBOP2016\BBCP.
\newblock \BBOQ Informed {M}onte {C}arlo {T}ree {S}earch for real-time strategy
  games\BBCQ\
\newblock In {\Bem IEEE Conference on Computational Intelligence and Games (CIG
  2016)}.

\bibitem[\protect\BCAY{Onta{\~n}{\'o}n\ \BBA\ Buro}{Onta{\~n}{\'o}n\ \BBA\
  Buro}{2015}]{ontanon2015adversarial}
Onta{\~n}{\'o}n, S.\BBACOMMA\  \BBA\ Buro, M. \BBOP2015\BBCP.
\newblock \BBOQ Adversarial hierarchical-task network planning for complex
  real-time games\BBCQ\
\newblock In {\Bem Proceedings of the 24th International Joint Conference on
  Artificial Intelligence}, \BPGS\ 1652--1658. AAAI Press.

\bibitem[\protect\BCAY{Onta{\~n}{\'o}n, Mishra, Sugandh,\ \BBA\
  Ram}{Onta{\~n}{\'o}n et~al.}{2007}]{ontanon2007case}
Onta{\~n}{\'o}n, S., Mishra, K., Sugandh, N., \BBA\ Ram, A. \BBOP2007\BBCP.
\newblock \BBOQ Case-based planning and execution for real-time strategy
  games\BBCQ\
\newblock In {\Bem Case-Based Reasoning Research and Development}, \BPGS\
  164--178. Springer.

\bibitem[\protect\BCAY{Onta{\~n}\'{o}n, Synnaeve, Uriarte, Richoux, Churchill,\
  \BBA\ Preuss}{Onta{\~n}\'{o}n et~al.}{2013}]{starcraft2013survey}
Onta{\~n}\'{o}n, S., Synnaeve, G., Uriarte, A., Richoux, F., Churchill, D.,
  \BBA\ Preuss, M. \BBOP2013\BBCP.
\newblock \BBOQ A survey of real-time strategy game {AI} research and
  competition in {StarCraft}\BBCQ\
\newblock {\Bem IEEE Transactions on Computational Intelligence and {AI} in
  Games (TCIAIG)}, {\Bem 5}, 1--19.

\bibitem[\protect\BCAY{Saffidine, Finnsson,\ \BBA\ Buro}{Saffidine
  et~al.}{2012}]{SaffidineFinnssonBuro2012AAAI}
Saffidine, A., Finnsson, H., \BBA\ Buro, M. \BBOP2012\BBCP.
\newblock \BBOQ Alpha-beta pruning for games with simultaneous moves\BBCQ\
\newblock In {\Bem Proceedings of the Association for the Advancement of
  Artificial Intelligence conference (AAAI 2012)}, Toronto, Canada. AAAI Press.

\bibitem[\protect\BCAY{Sharma, Holmes, Santamar{\'\i}a, Irani, Isbell~Jr,\
  \BBA\ Ram}{Sharma et~al.}{2007}]{sharma2007transfer}
Sharma, M., Holmes, M.~P., Santamar{\'\i}a, J.~C., Irani, A., Isbell~Jr, C.~L.,
  \BBA\ Ram, A. \BBOP2007\BBCP.
\newblock \BBOQ Transfer learning in real-time strategy games using hybrid
  {CBR/RL}\BBCQ\
\newblock In {\Bem Proceedings of the International Joint Conference on
  Artificial Intelligence (IJCAI 2007)}, \lowercase{\BVOL}~7, \BPGS\
  1041--1046.

\bibitem[\protect\BCAY{Shleyfman, Komenda,\ \BBA\ Domshlak}{Shleyfman
  et~al.}{2014}]{shleyfman2014combinatorial}
Shleyfman, A., Komenda, A., \BBA\ Domshlak, C. \BBOP2014\BBCP.
\newblock \BBOQ On combinatorial actions and {CMAB}s with linear side
  information\BBCQ\
\newblock In {\Bem Proceedings of the European Conference on Machine Learning
  (ECML 2014)}. Springer.

\bibitem[\protect\BCAY{Silver, Huang, Maddison, Guez, Sifre, van~den Driessche,
  Schrittwieser, Antonoglou, Panneershelvam,\ \BBA\ Lanctot}{Silver
  et~al.}{2016}]{silver2016mastering}
Silver, D., Huang, A., Maddison, C.~J., Guez, A., Sifre, L., van~den Driessche,
  G., Schrittwieser, J., Antonoglou, I., Panneershelvam, V., \BBA\ Lanctot, M.
  \BBOP2016\BBCP.
\newblock \BBOQ Mastering the game of {Go} with deep neural networks and tree
  search\BBCQ\
\newblock {\Bem Nature}, {\Bem 529\/}(7587), 484--489.

\bibitem[\protect\BCAY{Stanescu, Barriga,\ \BBA\ Buro}{Stanescu
  et~al.}{2014}]{stanescu2014hierarchical}
Stanescu, M., Barriga, N.~A., \BBA\ Buro, M. \BBOP2014\BBCP.
\newblock \BBOQ Hierarchical adversarial search applied to real-time strategy
  games\BBCQ\
\newblock In {\Bem Proceedings of the AAAI Artificial Intelligence and
  Interactive Digital Entertainment conference (AIIDE 2014)}.

\bibitem[\protect\BCAY{Stanescu, Barriga, Hess,\ \BBA\ Buro}{Stanescu
  et~al.}{2016}]{stanescu2016evaluating}
Stanescu, M., Barriga, N.~A., Hess, A., \BBA\ Buro, M. \BBOP2016\BBCP.
\newblock \BBOQ Evaluating real-time strategy game states using convolutional
  neural networks\BBCQ\
\newblock In {\Bem Proceedings of the IEEE Conference on Computational
  Intelligence in Games (CIG 2016)}.

\bibitem[\protect\BCAY{Tolpin\ \BBA\ Shimony}{Tolpin\ \BBA\
  Shimony}{2012}]{tolpin2012mcts}
Tolpin, D.\BBACOMMA\  \BBA\ Shimony, S.~E. \BBOP2012\BBCP.
\newblock \BBOQ {MCTS} based on simple regret\BBCQ\
\newblock In {\Bem Proceedings of the Association for the Advancement of
  Artificial Intelligence conference (AAAI 2012)}.

\bibitem[\protect\BCAY{Tromp\ \BBA\ Farneb{\"a}ck}{Tromp\ \BBA\
  Farneb{\"a}ck}{2006}]{tromp2006combinatorics}
Tromp, J.\BBACOMMA\  \BBA\ Farneb{\"a}ck, G. \BBOP2006\BBCP.
\newblock \BBOQ Combinatorics of {Go}\BBCQ\
\newblock In {\Bem International Conference on Computers and Games}, \BPGS\
  84--99. Springer.

\bibitem[\protect\BCAY{Uriarte\ \BBA\ Onta{\~n}{\'o}n}{Uriarte\ \BBA\
  Onta{\~n}{\'o}n}{2014}]{uriarte2014game}
Uriarte, A.\BBACOMMA\  \BBA\ Onta{\~n}{\'o}n, S. \BBOP2014\BBCP.
\newblock \BBOQ Game-tree search over high-level game states in {RTS}
  games\BBCQ\
\newblock In {\Bem Proceedings of the AAAI Artificial Intelligence and
  Interactive Digital Entertainment conference (AIIDE 2014)}.

\bibitem[\protect\BCAY{{Usunier}, {Synnaeve}, {Lin},\ \BBA\
  {Chintala}}{{Usunier} et~al.}{2016}]{2016arXiv160902993U}
{Usunier}, N., {Synnaeve}, G., {Lin}, Z., \BBA\ {Chintala}, S. \BBOP2016\BBCP.
\newblock \BBOQ {Episodic Exploration for Deep Deterministic Policies: An
  Application to StarCraft Micromanagement Tasks}\BBCQ\
\newblock {\Bem ArXiv e-prints}, {\Bem arXiv:1609.02993}.

\bibitem[\protect\BCAY{Weber\ \BBA\ Mateas}{Weber\ \BBA\
  Mateas}{2009}]{weber2009case}
Weber, B.~G.\BBACOMMA\  \BBA\ Mateas, M. \BBOP2009\BBCP.
\newblock \BBOQ Case-based reasoning for build order in real-time strategy
  games.\BBCQ\
\newblock In {\Bem Proceedings of the Artificial Intelligence and Interactive
  Digital Entertainment Conference (AIIDE 2009)}.

\bibitem[\protect\BCAY{Wender\ \BBA\ Watson}{Wender\ \BBA\
  Watson}{2012}]{wender2012applying}
Wender, S.\BBACOMMA\  \BBA\ Watson, I. \BBOP2012\BBCP.
\newblock \BBOQ Applying reinforcement learning to small scale combat in the
  real-time strategy game {StarCraft}: {Broodwar}\BBCQ\
\newblock In {\Bem Computational Intelligence and Games (CIG), 2012 IEEE
  Conference on}, \BPGS\ 402--408. IEEE.

\end{thebibliography}





\end{document}